\documentclass[11pt]{article}

\usepackage[margin=1in]{geometry}
\usepackage[T1]{fontenc}
\usepackage[utf8]{inputenc}
\usepackage{graphicx}
\usepackage{booktabs}
\usepackage{amsmath}
\usepackage{microtype}
\usepackage{caption}
\usepackage[section]{placeins}

\usepackage[hidelinks]{hyperref}
\usepackage{natbib}
\graphicspath{{figures/}}
\title{Explore or Confirm?\\
LLM Agents Misallocate Measurements Under Noise}

\author{
  Abdul Karim\thanks{Corresponding author.} \\ Xstendyl \\ \texttt{abdul.karim@xstendyl.com}
  \and
  Matt Keon \\ Xstendyl \\ \texttt{matt.keon@xstendyl.com}
}
\date{}

\newcommand{\nShareOpus}{74.9}%
\newcommand{\nShareOpusci}{[73.5, 76.3]}%
\newcommand{\nShareOpusint}{75}%
\newcommand{\nNonlostOpus}{77.4}%
\newcommand{\nNonlostOpusci}{[76.1, 78.7]}%
\newcommand{\nPickOpus}{7.621}%
\newcommand{\nPickOpusci}{[7.543, 7.696]}%
\newcommand{\nExplOpus}{1.02}%
\newcommand{\nExplOpusci}{[0.96, 1.09]}%
\newcommand{\nSelOpus}{0.12}%
\newcommand{\nSelOpusci}{[0.08, 0.16]}%
\newcommand{\nShareFlash}{94.0}%
\newcommand{\nShareFlashci}{[93.1, 94.8]}%
\newcommand{\nShareFlashint}{94}%
\newcommand{\nNonlostFlash}{94.6}%
\newcommand{\nNonlostFlashci}{[93.8, 95.3]}%
\newcommand{\nPickFlash}{7.701}%
\newcommand{\nPickFlashci}{[7.680, 7.720]}%
\newcommand{\nExplFlash}{1.04}%
\newcommand{\nExplFlashci}{[1.03, 1.06]}%
\newcommand{\nSelFlash}{0.02}%
\newcommand{\nSelFlashci}{[0.00, 0.03]}%
\newcommand{\nShareDeep}{91.0}%
\newcommand{\nShareDeepci}{[90.3, 91.6]}%
\newcommand{\nNonlostDeep}{91.2}%
\newcommand{\nNonlostDeepci}{[90.7, 91.8]}%
\newcommand{\nPickDeep}{7.760}%
\newcommand{\nPickDeepci}{[7.716, 7.799]}%
\newcommand{\nExplDeep}{0.94}%
\newcommand{\nExplDeepci}{[0.90, 0.98]}%
\newcommand{\nSelDeep}{0.06}%
\newcommand{\nSelDeepci}{[0.04, 0.08]}%
\newcommand{\nShareSol}{44.1}%
\newcommand{\nShareSolci}{[41.5, 46.7]}%
\newcommand{\nShareSolint}{44}%
\newcommand{\nNonlostSol}{45.7}%
\newcommand{\nNonlostSolci}{[43.1, 48.4]}%
\newcommand{\nPickSol}{7.672}%
\newcommand{\nPickSolci}{[7.582, 7.764]}%
\newcommand{\nExplSol}{0.93}%
\newcommand{\nExplSolci}{[0.84, 1.01]}%
\newcommand{\nSelSol}{0.16}%
\newcommand{\nSelSolci}{[0.11, 0.22]}%
\newcommand{\nSharePro}{14.0}%
\newcommand{\nShareProci}{[12.6, 15.3]}%
\newcommand{\nShareProint}{14}%
\newcommand{\nNonlostPro}{17.5}%
\newcommand{\nNonlostProci}{[15.9, 19.1]}%
\newcommand{\nPickPro}{7.688}%
\newcommand{\nPickProci}{[7.580, 7.802]}%
\newcommand{\nExplPro}{0.70}%
\newcommand{\nExplProci}{[0.60, 0.79]}%
\newcommand{\nSelPro}{0.38}%
\newcommand{\nSelProci}{[0.30, 0.46]}%
\newcommand{\nGuessShareInt}{24}%
\newcommand{\nBandHiInt}{38}%
\newcommand{\nPqNOpus}{30}%
\newcommand{\nPqShareOpus}{62.8}%
\newcommand{\nPqShareOpusci}{[61.0, 64.6]}%
\newcommand{\nPqPickOpus}{1.893}%
\newcommand{\nPqPickOpusci}{[1.753, 2.025]}%
\newcommand{\nPqLeadOpus}{$-$0.16}%
\newcommand{\nPqLeadOpusci}{[$-$0.30, $-$0.02]}%
\newcommand{\nPqShareFlash}{63.7}%
\newcommand{\nPqShareFlashci}{[60.6, 66.6]}%
\newcommand{\nPqPickFlash}{2.130}%
\newcommand{\nPqPickFlashci}{[2.050, 2.184]}%
\newcommand{\nPqLeadFlash}{$+$0.08}%
\newcommand{\nPqLeadFlashci}{[$-$0.01, $+$0.15]}%
\newcommand{\nPqShareDeep}{89.9}%
\newcommand{\nPqShareDeepci}{[87.5, 91.9]}%
\newcommand{\nPqPickDeep}{0.829}%
\newcommand{\nPqPickDeepci}{[0.757, 0.942]}%
\newcommand{\nPqLeadDeep}{$-$1.22}%
\newcommand{\nPqLeadDeepci}{[$-$1.31, $-$1.10]}%
\newcommand{\nPqShareSol}{23.9}%
\newcommand{\nPqShareSolci}{[20.8, 27.3]}%
\newcommand{\nPqPickSol}{2.368}%
\newcommand{\nPqPickSolci}{[2.155, 2.582]}%
\newcommand{\nPqLeadSol}{$+$0.32}%
\newcommand{\nPqLeadSolci}{[$+$0.10, $+$0.54]}%
\newcommand{\nPqTopSol}{14}%
\newcommand{\nPqSharePro}{14.7}%
\newcommand{\nPqShareProci}{[13.1, 16.3]}%
\newcommand{\nPqPickPro}{1.725}%
\newcommand{\nPqPickProci}{[1.601, 1.857]}%
\newcommand{\nPqLeadPro}{$-$0.33}%
\newcommand{\nPqLeadProci}{[$-$0.46, $-$0.19]}%
\newcommand{\nPqOamShareOpus}{65.3}%
\newcommand{\nPqOamShareOpusci}{[63.0, 67.6]}%
\newcommand{\nPqOamPickOpus}{1.89}%
\newcommand{\nPqOamShareFlash}{61.7}%
\newcommand{\nPqOamShareFlashci}{[58.7, 64.5]}%
\newcommand{\nPqOamPickFlash}{2.12}%
\newcommand{\nPqOamShareDeep}{91.4}%
\newcommand{\nPqOamShareDeepci}{[90.3, 92.3]}%
\newcommand{\nPqOamPickDeep}{0.81}%
\newcommand{\nPqGuessShare}{1.03}%
\newcommand{\nPqGuessPick}{2.050}%
\newcommand{\nPqSpend}{20.89}%
\newcommand{\nGbNOpus}{60}%
\newcommand{\nPqFlashLostRounds}{3.7}%
\newcommand{\nGridOneReadingPro}{78}%
\newcommand{\nGridOneReadingInf}{2}%
\newcommand{\nGridOneReadingGuess}{2}%
\newcommand{\nPairGbOpus}{$-$1.41}%
\newcommand{\nPairGbOpusci}{[$-$1.92, $-$0.89]}%
\newcommand{\nPairPqOpus}{$-$0.26}%
\newcommand{\nPairPqOpusci}{[$-$0.60, $+$0.07]}%
\newcommand{\nPairGbFlash}{$-$5.07}%
\newcommand{\nPairGbFlashci}{[$-$5.47, $-$4.68]}%
\newcommand{\nPairPqFlash}{$-$0.02}%
\newcommand{\nPairPqFlashci}{[$-$0.35, $+$0.28]}%
\newcommand{\nPairGbDeep}{$-$5.35}%
\newcommand{\nPairGbDeepci}{[$-$5.75, $-$4.94]}%
\newcommand{\nPairPqDeep}{$-$1.32}%
\newcommand{\nPairPqDeepci}{[$-$1.68, $-$0.98]}%
\newcommand{\nPairGbSol}{$-$0.06}%
\newcommand{\nPairGbSolci}{[$-$0.65, $+$0.52]}%
\newcommand{\nPairPqSol}{$+$0.22}%
\newcommand{\nPairPqSolci}{[$-$0.21, $+$0.62]}%
\newcommand{\nPairGbPro}{$-$1.52}%
\newcommand{\nPairGbProci}{[$-$2.44, $-$0.55]}%
\newcommand{\nPairPqPro}{$-$0.42}%
\newcommand{\nPairPqProci}{[$-$0.78, $-$0.10]}%
\newcommand{\nPqSolLeadReg}{$+$0.32}%
\newcommand{\nPqSolLeadRegci}{[$+$0.10, $+$0.54]}%
\newcommand{\nPqSolLeadNom}{$+$0.32}%
\newcommand{\nPqSolLeadNomci}{[$-$0.09, $+$0.55]}%
\newcommand{\nPqSolLeadOnce}{$+$0.20}%
\newcommand{\nPqSolLeadOnceci}{[$-$0.14, $+$0.57]}%
\newcommand{\nPqSolLeadOnceNom}{$+$0.20}%
\newcommand{\nPqSolLeadOnceNomci}{[$-$0.16, $+$0.61]}%
\newcommand{\nPqSolDistinctNominees}{13}%
\newcommand{\nBandHi}{37.61}%
\newcommand{\nProOneReading}{78}%
\newcommand{\nProHeavyCalls}{35}%
\newcommand{\nProMaxReasoning}{30{,}721}%
\newcommand{\nAoneProKthree}{$+$0.088}%
\newcommand{\nAoneProKthreeci}{[$+$0.011, $+$0.171]}%
\newcommand{\nAtwoProKthree}{$+$0.100}%
\newcommand{\nAtwoProKthreeci}{[$+$0.013, $+$0.191]}%
\newcommand{\nAoneProRecoveredKthree}{23}%
\newcommand{\nAtwoProRecoveredOwn}{27}%
\newcommand{\nSolGbLeadReg}{$+$0.28}%
\newcommand{\nSolGbLeadRegci}{[$+$0.06, $+$0.53]}%
\newcommand{\nSolGbLeadClu}{$+$0.28}%
\newcommand{\nSolGbLeadCluci}{[$-$0.01, $+$0.74]}%
\newcommand{\nSolGbLeadCol}{$+$0.33}%
\newcommand{\nSolGbLeadColci}{[$-$0.01, $+$0.71]}%
\newcommand{\nEpisodesPerSec}{159}%
\newcommand{\nSplitFlashExpl}{$+$0.17}%
\newcommand{\nSplitFlashExplci}{[$+$0.11, $+$0.23]}%
\newcommand{\nSplitFlashSel}{$+$0.13}%
\newcommand{\nSplitFlashSelci}{[$+$0.10, $+$0.17]}%
\newcommand{\nSplitDeepExpl}{$+$0.08}%
\newcommand{\nSplitDeepExplci}{[$+$0.01, $+$0.16]}%
\newcommand{\nSplitDeepSel}{$+$0.10}%
\newcommand{\nSplitDeepSelci}{[$+$0.05, $+$0.14]}%
\newcommand{\nSplitOpusSel}{$+$0.09}%
\newcommand{\nSplitOpusSelci}{[$+$0.03, $+$0.16]}%
\newcommand{\nAoneDeepKone}{$+$0.035}%
\newcommand{\nAoneDeepKoneci}{[$+$0.016, $+$0.055]}%
\newcommand{\nAoneFlashKone}{$+$0.014}%
\newcommand{\nAoneFlashKoneci}{[$+$0.002, $+$0.032]}%
\newcommand{\nAoneProKfour}{$+$0.090}%
\newcommand{\nAoneProKfourci}{[$-$0.004, $+$0.190]}%
\newcommand{\nAoneProRepick}{$+$0.004}%
\newcommand{\nAoneProRepickci}{[$-$0.028, $+$0.036]}%
\newcommand{\nAoneProShareAfter}{18.9}%
\newcommand{\nAtwoSolKone}{$-$0.041}%
\newcommand{\nAtwoSolKoneci}{[$-$0.100, $+$0.011]}%
\newcommand{\nSolGbMenuClu}{$-$0.25}%
\newcommand{\nSolGbMenuCluci}{[$-$1.10, $+$0.41]}%
\newcommand{\nSolGbMenuCol}{$-$0.29}%
\newcommand{\nSolGbMenuColci}{[$-$1.14, $+$0.54]}%
\newcommand{\nFsrmTrue}{6.275}%
\newcommand{\nGemSpend}{71.55}%
\newcommand{\nSolSpend}{20.39}%
\newcommand{\nGridSelLow}{0.18}%
\newcommand{\nGridSelAssay}{0.36}%
\newcommand{\nGridSelHigh}{0.54}%
\newcommand{\nGridLostKzero}{23}%
\newcommand{\nGridLostKone}{12}%
\newcommand{\nPhoqTimesLow}{4}%
\newcommand{\nPhoqTimesHigh}{8}%
\newcommand{\nGbTimesHigh}{8}%
\newcommand{\nGbHighKone}{4.20}%
\newcommand{\nGbHighKoneci}{[4.11, 4.29]}%
\newcommand{\nGbHighKzero}{3.60}%
\newcommand{\nGbHighKzeroci}{[3.53, 3.67]}%
\newcommand{\nGridRfKone}{7.95}%
\newcommand{\nGridRfKoneci}{[7.93, 7.97]}%
\newcommand{\nGridRfKzero}{7.79}%
\newcommand{\nGridRfKzeroci}{[7.77, 7.82]}%
\newcommand{\nBandNonlostLo}{29.80}%
\newcommand{\nBandNonlostHi}{44.10}%
\newcommand{\nPanelcOpus}{69}%
\newcommand{\nPanelcFlash}{82}%
\newcommand{\nGbDistinctFlash}{3.9}%
\newcommand{\nGbDistinctDeep}{2.4}%
\newcommand{\nGbShareOpus}{61.7}%
\newcommand{\nGbDistinctOpus}{19.8}%
\newcommand{\nGbDistinctSol}{45.8}%
\newcommand{\nGbDistinctPro}{46.9}%
\newcommand{\nGbExplPro}{3.76}%
\newcommand{\nGbSelPro}{0.10}%
\newcommand{\nGbBestShare}{0.9}%
\newcommand{\nPolRandGrid}{0.40}%
\newcommand{\nPolRandSpike}{0.09}%
\newcommand{\nPolRandSpikeci}{[0.05, 0.13]}%
\newcommand{\nPolSpikeHeavy}{$+$0.03}%
\newcommand{\nPolSpikeHeavyci}{[$-$0.01, $+$0.07]}%
\newcommand{\nPolEntRandPq}{0.50}%
\newcommand{\nPolEntRandPqci}{[0.43, 0.56]}%
\newcommand{\nPolShare}{9.3}%
\newcommand{\nPolParentTopThree}{61}%
\newcommand{\nPolLeadRetest}{0.3}%
\newcommand{\nPolRetry}{62}%
\newcommand{\nPolSameRound}{15}%
\newcommand{\nGuessShare}{23.78}%
\newcommand{\nInformedPick}{8.00}%
\newcommand{\nGbViable}{7.2}%
\newcommand{\nSliceNbrs}{38}%
\newcommand{\nSliceNbrsViable}{20}%
\newcommand{\nSliceViablePct}{23}%
\newcommand{\nStartSliceViable}{16}%
\newcommand{\nStartSliceCells}{400}%
\newcommand{\nStartRowCol}{38}%
\newcommand{\nStartRowColViable}{4}%
\newcommand{\nStartOffSliceViable}{20}%
\newcommand{\nStartsRandom}{50}%
\newcommand{\nStartGuessShareMinGrid}{23.4}%
\newcommand{\nStartGuessShareMaxGrid}{23.8}%
\newcommand{\nStartMgShareMinGrid}{37.0}%
\newcommand{\nStartMgShareMaxGrid}{37.9}%
\newcommand{\nStartMgPickLostMaxGrid}{0.059}%
\newcommand{\nStartMgPickLostMedGrid}{0.008}%
\newcommand{\nStartSameKGbMg}{49}%
\newcommand{\nStartSameKPqMg}{50}%
\newcommand{\nTopSpreadGrid}{0.32}%
\newcommand{\nTopSpreadGb}{2.95}%
\newcommand{\nTopSpreadPq}{6.80}%
\newcommand{\nNbrPer}{76}%
\newcommand{\nNbrViableGrid}{25.8}%
\newcommand{\nStartNbrsGrid}{24}%
\newcommand{\nNbrViableGb}{45.9}%
\newcommand{\nNbrViablePq}{37.6}%

\begin{document}
\maketitle

\begin{abstract}
As model reasoning gets cheaper, the physical experiment becomes the slow and costly step in discovery. Each experiment returns a noisy reading: a high value can mean a good candidate or a lucky draw. With a fixed budget, an AI agent must keep choosing between exploring a new candidate and confirming a promising one by re-testing it. We measure that choice in Wellocate, a simulated assay campaign with known true values. It is built on a generated landscape and on two real protein datasets, GB1 and PhoQ, scrambled so no model can answer from memory. We test five language models as minimal agents, with no tools and no memory beyond the history in each prompt. They misallocate in both directions. Three over-confirm, spending most of the budget re-testing. On the generated landscape one under-confirms, trusting single readings, and there every model ends with a worse final pick than a basic scripted strategy. Each failure has its own cause. The over-confirmers run short of new candidates worth trying: a list of untried candidates cuts their re-testing and, on real protein data, makes two of them pick much better. A running summary of their readings or a description of the landscape changes it by at most five percentage points. For the under-confirmer, re-testing its top candidates in a final round, replayed on its own campaigns, recovers only about a quarter of what it loses by picking the wrong candidate. This suggests confirmation has to run through the whole campaign. We release Wellocate as a benchmark and reinforcement-learning environment with frozen splits and scripted baselines. With it, anyone can measure how an agent splits its budget before trusting it with real experiments, and train agents to split it better. Code, data and the environment are available at
\url{https://github.com/Xstendyl/wellocate}.
\end{abstract}

\section{Introduction}
\label{sec:intro}

As model reasoning gets cheaper and faster, the physical experiment becomes the bottleneck in discovery. Assays,
syntheses and fabrication runs keep their cost and their wall-clock time, and each one returns a noisy reading. So the
value of an AI agent in science depends more and more on how it spends a fixed experiment budget. In a protein or drug
discovery lab, each round someone, or now some agent, decides which variants go to the assay. Each slot gets a new candidate or another reading of one that looked good. This is the choice we study in this paper: explore or confirm.

A reading that comes back high may be a good candidate or a lucky draw. Re-testing tells which, and spends a slot that
could have tried something new. A campaign is judged like a batting average: what counts is the useful output of the
whole budget, and a test that finds nothing costs only its slot. So a re-test earns its slot only if it is worth more
than the new candidate it displaces. How often to repeat a measurement is an old question in experimental design. Sequential design
methods choose at each step between a new point and a repeat of an old one, and repeats tend to pay off where the
noise is large \citep{binois2019replication}. Batch Bayesian optimization goes a step further and picks both the
candidates in each batch and how many times to measure each one \citep{dai2023replicable}. Self-driving labs
already choose experiments by algorithm \citep{rapp2024selfdriving,tom2024selfdriving}, and language-model agents
now design experiments \citep{boiko2023coscientist,roohani2024biodiscovery} and proteins \citep{biodesignbench2026}. To our knowledge, how such agents split a noisy budget has
not been measured against known true values.

We build Wellocate to measure this choice: a simulated assay campaign in which every candidate's true value is
known. It runs on a generated landscape, which we call the grid, and on two real protein landscapes, GB1 and PhoQ
\citep{wu2016adaptation,podgornaia2015phoq}. We chose proteins because protein assays are exactly this kind of slow, costly and noisy step. Their amino-acid names are scrambled so no model can answer from memory. We run five language models on
the grid, GB1 and PhoQ. Each runs as a minimal agent with no tools and no memory beyond the history of its orders and readings, which it is shown each round. What we measure is thus the model's own allocation.

\begin{figure}[!htb]
\centering
\includegraphics[width=\textwidth]{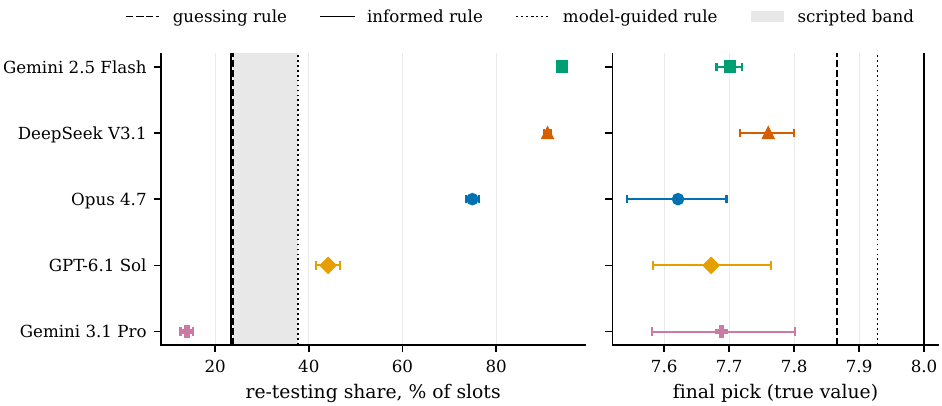}
\caption{Five models on the grid. Left: the re-testing share,
the fraction of used slots spent on a candidate the agent had already ordered, with 95\% intervals. Right: the true value of
each model's final pick. The lines are the three scripted rules (Section~\ref{sec:env}), drawn the same way in both panels. The guessing rule is dashed, the model-guided rule dotted, and the informed rule, which is also told which candidates are viable, solid. The grey band is the scripted band, from the guessing rule's share to the model-guided rule's. The informed
rule re-tests about as often as the guessing rule, so their lines overlap on the left. On the grid, Opus 4.7,
Gemini 2.5 Flash and DeepSeek V3.1 re-test far above the band, GPT-6.1 Sol just above it and Gemini 3.1 Pro below it. Every model's pick falls short of all three scripted rules. The figure and its band are for the grid only; on GB1 and PhoQ the best scripted rule re-tests at most 1\% of its slots (Table~\ref{tab:landscapes}).}
\label{fig:headline}
\end{figure}

To say a model re-tests too much or too little, we need a yardstick. We use three simple fixed strategies, the \emph{scripted rules}, that face
the same budget and noise as the models. The \emph{guessing rule} tries small changes to its best candidate, guessing which ones are viable. Most candidates do not work, and an order for one that does not is rejected. The rule re-tests its best a fixed number of times each round, set for each landscape. The \emph{model-guided rule} is smarter about what to try. It fits a simple model to its own readings, estimating how much each letter at each position adds to a candidate's value. It then orders the new candidates that model predicts will be best. It re-tests one leading candidate each round and
spends its last three rounds re-testing its four best. The \emph{informed rule} plays like the guessing rule and is also told which candidates are viable; it still has to measure how good they are. It is
never rejected, so it loses no slots to wrong guesses, an advantage the models do not have. On the grid and PhoQ it
makes the best final pick. On the grid the guessing rule
spends about \nGuessShareInt\% of its slots on re-tests and the model-guided rule about \nBandHiInt\%, with a better final pick.
Together these two mark the \emph{scripted band} of re-testing on the grid. Against that band the models fall into three groups (Figure~\ref{fig:headline}). Three
over-confirm: Opus 4.7, Gemini 2.5 Flash and DeepSeek V3.1 spend between \nShareOpusint\% and \nShareFlashint\% of
their slots re-testing. One under-confirms: Gemini 3.1 Pro re-tests only \nShareProint\% and often picks its
winner on a single reading. GPT-6.1 Sol, at \nShareSolint\%, sits just above the band. On the grid, every model picks a worse final
candidate than the guessing rule, whether it re-tests too much or too little. On GB1 and PhoQ
the best scripted rule re-tests at most 1\% of its slots, and every model re-tests more than it does, Gemini 3.1 Pro
included (Section~\ref{sec:both}, Table~\ref{tab:landscapes}).

Each failure has its own cause. We give the agents three simple aids. They are a running summary of their own readings
(the ledger), a note on how the landscape is built, and a list of untried candidates (the Menu). The ledger and the landscape note change how much they re-test by at most five percentage points. The Menu cuts re-testing in all three over-confirming models. On GB1,
without the Menu, Gemini 2.5 Flash and DeepSeek V3.1 test on average only \nGbDistinctFlash{} and
\nGbDistinctDeep{} different candidates in a 60-measurement campaign. Given the Menu, their final picks close most of
the gap to the guessing rule. These models re-test because they run short of new candidates worth trying, so the fix
is to supply candidates. Gemini 3.1 Pro trusts single readings. To test the simplest fix,
we replay its campaigns and replace its last round of orders with re-tests of its top candidates. This recovers only
about a quarter of what it loses by picking the wrong candidate, which suggests confirmation has to run through the
whole campaign. This paper contributes:
\begin{itemize}
\item A measurement of how five models split a noisy budget (Section~\ref{sec:both}). Their loss divides into
exploration loss, from never measuring better candidates, and selection loss, from picking the wrong one among those measured.
\item Evidence on what drives each failure. The over-confirmers run out of new candidates worth trying. The Menu cuts their re-testing, and a summary of their readings or a note on the landscape changes it by at most five percentage points
(Section~\ref{sec:why}). The under-confirmer picks winners on one noisy reading, and a final round of re-tests
recovers only about a quarter of that loss. This suggests confirmation must run through the whole campaign
(Section~\ref{sec:under}). Section~\ref{sec:discussion} turns this into steps a lab can follow.
\item Wellocate itself, released as a benchmark and reinforcement-learning environment with a protein track built from real data,
frozen splits and scripted baselines (Section~\ref{sec:bench}). With it, anyone can measure how an agent splits its budget before trusting it with real experiments, and train agents to split it better.
\end{itemize}

\section{Wellocate: a simulated assay campaign}
\label{sec:env}

\begin{figure}[tbp]
\centering
\includegraphics[width=\textwidth]{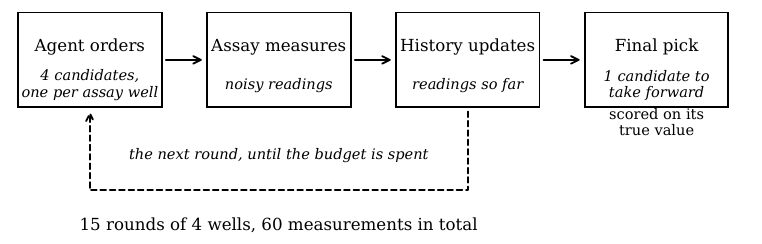}
\caption{One Wellocate campaign, with the lab counterpart of each part. Each round the agent orders four candidates,
one per assay well. The simulated assay returns a noisy reading for each, and the agent sees every reading so far
before ordering again. Orders for non-viable candidates are rejected and replaced, and about 15\% of orders for viable candidates
fail. After 15 rounds the agent nominates the candidate to take forward, scored on its true value.}
\label{fig:env}
\end{figure}

\emph{Wellocate} turns an assay campaign into an episode with a score (Figure~\ref{fig:env}). We copy its shape
from two real protein experiments, deep mutational scans that set out to measure every variant at four positions of GB1 and of
PhoQ \citep{wu2016adaptation,podgornaia2015phoq}. A candidate is a string of four letters, one per position, from a
20-letter alphabet, one letter for each amino acid, so the space holds 160{,}000 candidates. Each candidate has a hidden
true value, its quality, which the agent never sees.

\begin{figure}[tbp]
\centering
\includegraphics[width=\textwidth]{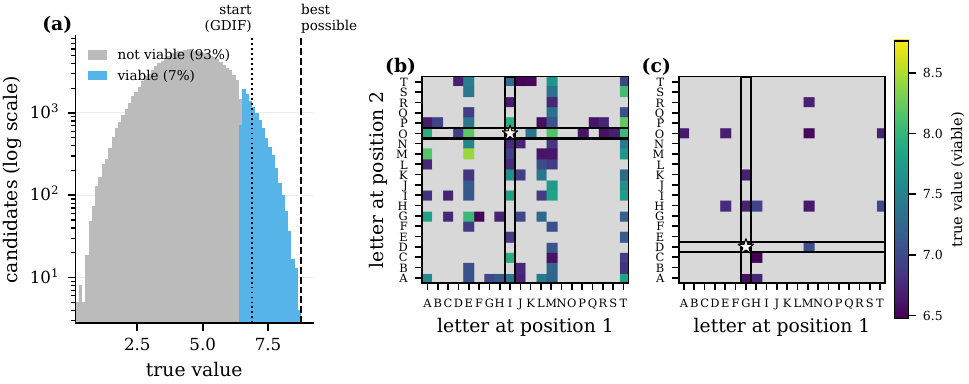}
\caption{The grid. (a) True value of all 160{,}000 candidates. A value averages four random position scores, so
most candidates sit near the middle and very high or very low values are rare. The vertical axis is on a log scale
so that the rare ones show. Only the top 7\% (11{,}199) are viable, shown in
blue; an order for any other candidate is rejected. Every campaign starts from \texttt{GDIF}, the median viable
candidate (6.87), and the best possible is 8.762. (b, c) A candidate has four letters, so each grid shows one
slice: letters 3 and 4 are fixed and letters 1 and 2 vary, 400 candidates per slice. Grey cells are not viable.
Changing only letter 1 moves along the star's row, and changing only letter 2 moves along its column. The outlined row and column thus hold the \nSliceNbrs{} candidates one letter away that a slice can show. (b) The slice through
the best candidate, \texttt{IOJO}, with J and O fixed: \nSliceViablePct\% of the slice is viable, and
\nSliceNbrsViable{} of the star's \nSliceNbrs{} outlined neighbours. This is the richest region of the grid. (c) The
slice through the start, \texttt{GDIF}, with I and F fixed: \nStartSliceViable{} of \nStartSliceCells{} cells are
viable, and \nStartRowColViable{} of the star's \nStartRowCol{} outlined neighbours. \nStartOffSliceViable{} of the
start's \nStartNbrsGrid{} viable neighbours change letter 3 or 4 and so lie off this slice. The search problem is
getting from (c) to (b).}
\label{fig:gridland}
\end{figure}

The true values come from a landscape, a fixed table that gives every candidate its value. We use three. One is
synthetic: a generated NK landscape \citep{kauffman1989nk} that we call the grid (Figure~\ref{fig:gridland}). Each of
its four positions gets a score between 0 and 1 from a fixed random table, and the score depends on that position's
letter and the next position's letter. A candidate's true value is the average of its four scores. The main study runs on the grid because every true value is known exactly and no model can
have seen its optimum. Two are real: the published landscapes of GB1 \citep{wu2016adaptation} and PhoQ \citep{podgornaia2015phoq}. We run
the same agents on them to check that what we find on the grid also holds on real biology. A model may have memorised
these, so their amino-acid names are replaced by the letters A to T through a random map at each position. Each variant keeps its
measured value under its new name. Two variants that differ by a single mutation in the real protein still
differ by a single letter after the renaming. The scrambled landscape is therefore exactly as hard to search as the real one, and a model cannot answer
from memory. Each landscape has its own best candidate (\texttt{IOJO} on the grid). We multiply every value in a
landscape by one constant so that its best candidate scores 8.762, the top reading in the GB1 data. This puts all three on one scale and changes no ranking. The two kinds of landscape also become the two
tracks of the benchmark we release (Section~\ref{sec:bench}). The \emph{grid} track generates fresh NK landscapes, and the \emph{protein} track uses fresh scrambles of GB1 and PhoQ.

Most candidates do not work at all. In a real protein, most random variants do not fold or
lose their function, and a search has to find the few that work. We call a candidate \emph{viable} if its true value is in the top 7\%
of its landscape, and use the same 7\% on all three landscapes (Figure~\ref{fig:gridland}a). The share is close to the share of GB1 variants that keep at least a tenth of the
wild type's fitness (\nGbViable\%). The agent is not told which candidates are viable. Viable candidates cluster, so the neighbours of a working candidate are a good place to look. Take any viable candidate
and its \nNbrPer{} neighbours, the candidates one letter away (19 other letters at each of 4 positions). On the grid, \nNbrViableGrid\% of those neighbours are viable on average, against 7\% of all candidates
(Figure~\ref{fig:gridland}b). The two real landscapes cluster more strongly (\nNbrViableGb\% on GB1,
\nNbrViablePq\% on PhoQ), so the grid does not make the search easier than real data would.

A campaign has 15 rounds of 4 slots, one slot per assay well, 60 measurements in all. Every campaign starts from the
same candidate, the median viable one (\texttt{GDIF} on the grid). The agent is told its value, as a lab starts
from a variant it already knows. One shared start gives every model the same search. Each round the agent orders
four candidates, and the environment, which holds every true value, handles each order. It rejects an order for a
non-viable candidate before the order uses a slot and asks the agent for a replacement. If three replacements in a
row are also rejected, the slot is lost. It fails about 15\% of orders for viable candidates at random, whatever
their value: nothing comes back, and the slot is still spent. Every other order returns a reading: the candidate's
true value plus noise with a standard deviation of 7.3\% of that value. The 7.3\% is the counting error of the
GB1 deep mutational scan \citep{wu2016adaptation}. On average it read each variant's DNA about 557 times before selection and 285 times after. Counting at those depths gives a relative error of about 7.3\% (Appendix~\ref{app:noise}). The environment keeps the record of
every order, rejection, failure and reading, and shows the agent the full record before each round. The agent may
order a candidate it has already measured. At the end it nominates one candidate, scored on its true value. The budget and the failure rate are design choices: the budget is sized for slow assays in which each measurement is costly.

Each piece of the campaign has a lab counterpart. A slot is an assay well. A failed order is a failed assay. A
non-viable candidate is a variant that does not fold or express, caught here before it uses a well. The noise is
replicate variability: the same variant measured twice gives two different numbers. The nominee is the variant taken
forward. Appendix~\ref{app:env} lists where the environment is simpler than a lab, and the direction of each
simplification.

A finished campaign leaves two things to score: how the agent spent its slots, and how good its final pick was. To score how the slots were spent, we use the \emph{re-testing share}, the fraction of used slots spent on a candidate the agent had already ordered. An agent that orders the same candidate in all four slots of a round re-tests three of them.
The share counts repeats over the whole campaign, whichever candidates they went to; how many readings lie behind
the final pick is a separate count, reported in Section~\ref{sec:both}. To score the final pick, we split its shortfall from the best candidate in the landscape into two parts that add up. Both are computed from true values, which the agent never sees. The \emph{exploration loss} is how far the best candidate the agent measured
falls short of the landscape's best; it is large when the search never reached the good candidates. The
\emph{selection loss} is how far the nominated candidate falls short of the best candidate the agent measured. It is large when the agent found a good candidate and then nominated a worse one. Selection loss comes from noise. The agent can rank
candidates only by their readings, and a candidate measured once can read well above its true value and look better
than one that is truly better. Re-testing averages this noise away and shrinks the selection loss. Each re-test also takes a slot from a new candidate, which can raise the exploration loss.

A re-testing share on its own does not say whether an agent re-tests too much or too little, so we need a yardstick.
We use three simple fixed strategies, together called the \emph{scripted rules}: the guessing rule, the informed
rule and the model-guided rule. All three work under the same budget, noise and order format as the agents. Each round, the \emph{guessing rule} re-tests its current best a fixed number of times. It fills the other slots
with untried neighbours of it, candidates one letter away, guessing which ones are viable, as the agents must. The \emph{informed rule} does the same and is also told which candidates are viable, so it is never rejected; it still has to measure how good they are. For these two rules we set the number of re-tests per round after the runs, to the one with the best mean true pick on each landscape. The agents get no such tuning. The \emph{model-guided rule} learns from its own
readings only. It scores each letter at each position by the mean reading of the measured candidates that carry it. It adds these scores to rank untried one- and two-letter changes of its current best. It re-tests one leading
candidate each round and spends its last three rounds re-testing its four best. Appendix~\ref{app:refs} gives the
settings of all three.

Against this yardstick we run five language models: Opus 4.7, Gemini 2.5 Flash, DeepSeek V3.1, GPT-6.1 Sol and
Gemini 3.1 Pro, each as a minimal agent. Every round it receives one prompt holding the task and the full history of its orders and readings, and it replies with its orders. That history, written by the environment, is its only memory of earlier rounds. The model itself carries nothing from one call to the next, and it has no tools or scaffold. We chose this on purpose, so that what we measure is the model's own allocation. 

\section{Results}
\label{sec:res}

\subsection{Misallocation in both directions}
\label{sec:both}

The five models, run under a registered design (Appendix~\ref{app:amendments}), misallocate their measurements in both directions.

The scripted rules set the yardstick (Table~\ref{tab:landscapes}). On the grid the guessing and model-guided rules
re-test \nGuessShare\% to \nBandHi\% of their slots, the \emph{scripted band}. The best final pick comes from the
informed rule, which is told which candidates are viable: a candidate worth \nInformedPick. On the two proteins, at the noise of the real assay (\emph{assay noise}), the
best scripted rule hardly re-tests at all: \nGbBestShare\% of slots on GB1 and none on PhoQ. With more noise,
re-testing pays there too (Section~\ref{sec:bench}).

Three models, Opus 4.7, Gemini 2.5 Flash and DeepSeek V3.1, over-confirm. On the grid they spend
\nShareOpusint\% to \nShareFlashint\% of their slots re-testing, far above the scripted band (Figure~\ref{fig:headline}). With every reading exact they still re-test half to nearly all of their slots, far more than exact readings
need (Appendix~\ref{app:grid}). Gemini 3.1 Pro under-confirms. It re-tests only \nShareProint\% of its slots, below the
band, so on the grid its pick is usually a candidate it measured only once. GPT-6.1 Sol sits just above the band.
On GB1 and PhoQ the best scripted rule re-tests at most 1\% of its slots. Some models re-test less there than on the grid, and every model still re-tests more than that rule, Gemini 3.1 Pro included (Table~\ref{tab:landscapes}). On the grid every model's final pick falls below the guessing rule's 7.87, and re-testing less did
not help Gemini 3.1 Pro or Sol pick better (Figure~\ref{fig:headline}). On GB1 Sol's lead over the guessing rule is uncertain (Appendix~\ref{app:atwo}). On PhoQ, a hard landscape for every strategy, three models are about level with it, and DeepSeek V3.1 and Gemini 3.1 Pro pick worse (Appendix~\ref{app:phoq}).

\begin{table}[tbp]
\centering
\small
\caption{Re-testing share and true value of the final pick on each landscape, averaged over each model's campaigns. The three scripted rules are shown at their best setting.}
\label{tab:landscapes}
\begin{tabular}{lrrrrrr}
\toprule
 & \multicolumn{2}{c}{grid} & \multicolumn{2}{c}{GB1} & \multicolumn{2}{c}{PhoQ} \\
\cmidrule(lr){2-3}\cmidrule(lr){4-5}\cmidrule(lr){6-7}
 & share (\%) & pick & share (\%) & pick & share (\%) & pick \\
\midrule
Opus 4.7 & 74.9 & 7.62 & 63.3 & 4.78 & 62.8 & 1.89 \\
Gemini 2.5 Flash & 94.0 & 7.70 & 91.9 & 1.12 & 63.7 & 2.13 \\
DeepSeek V3.1 & 91.0 & 7.76 & 96.0 & 0.84 & 89.9 & 0.83 \\
GPT-6.1 Sol & 44.1 & 7.67 & 19.4 & 6.36 & 23.9 & 2.37 \\
Gemini 3.1 Pro & 14.0 & 7.69 & 11.1 & 4.90 & 14.7 & 1.72 \\
\midrule
guessing rule & 23.8 & 7.87 & 0.9 & \textbf{6.09} & 1.0 & 2.05 \\
informed rule & 23.3 & \textbf{8.00} & 0.0 & 6.02 & 0.0 & \textbf{2.63} \\
model-guided rule & 37.6 & 7.93 & 22.6 & 5.61 & 25.3 & 1.89 \\
\bottomrule
\end{tabular}
\end{table}

\begin{figure}[tbp]
\centering
\includegraphics[width=0.68\textwidth]{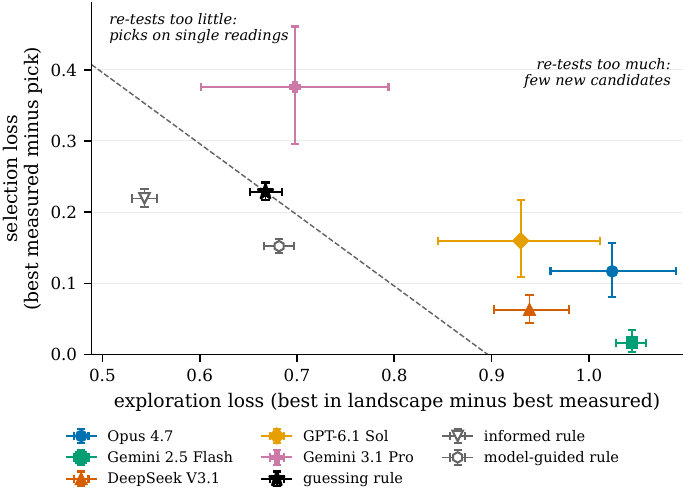}
\caption{Where each model's loss comes from, with 95\% intervals. Exploration loss: how far the best candidate
measured falls short of the landscape's best. Selection loss: how far the pick falls short of the best candidate
measured. The dashed line joins points with the guessing rule's total loss. The informed and model-guided rules are shown for reference; the informed rule is told which candidates are viable. Opus 4.7, Gemini 2.5 Flash and DeepSeek V3.1 measure too few candidates and pick well among them; Gemini 3.1 Pro measures as widely as the guessing rule and picks poorly.}
\label{fig:loss}
\end{figure}

The over-confirmers lose through exploration (Figure~\ref{fig:loss}). The best candidate they ever measured falls \nExplDeep{} to \nExplFlash{} short of the landscape's best, against 0.67 for the guessing rule. Among what they tested they pick almost perfectly. Gemini 3.1 Pro loses through selection. It explores about as well as the guessing rule and loses \nSelPro{} at
selection, against 0.23 for the guessing rule, because it often picks a candidate it measured only once. Sol loses about as much as the over-confirmers through exploration and less than the
guessing rule at selection.

\subsection{Over-confirmation}
\label{sec:why}

\begin{figure}[!t]
\centering
\includegraphics[width=\textwidth]{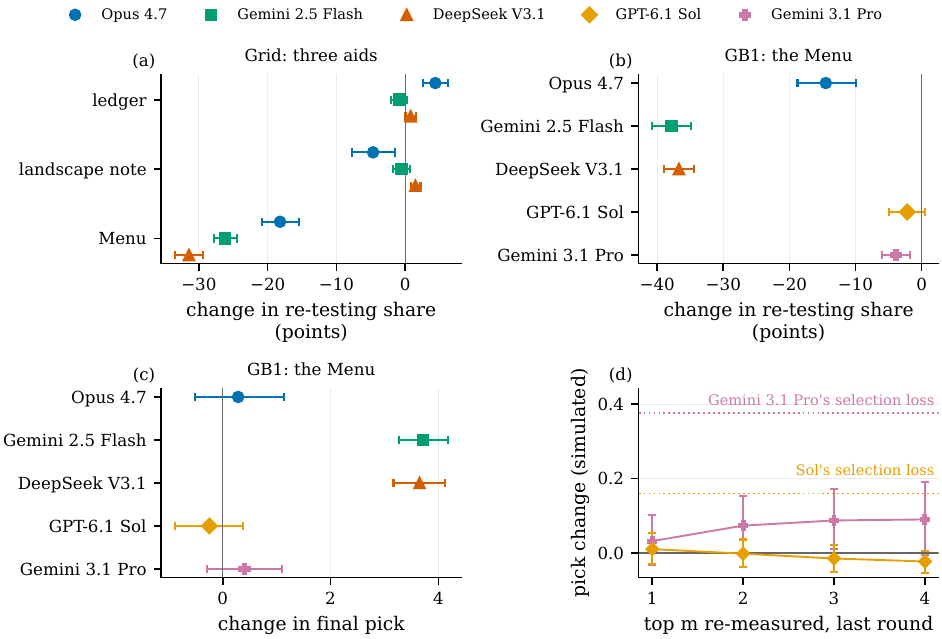}
\caption{What we tested for each failure, each against the same model's control on the same seeds, with 95\% intervals.
\emph{Over-confirmation.} (a) The three aids on the grid, given to the three over-confirming models: change in re-testing
share. (b, c) The Menu on scrambled GB1, given to all five models: change in re-testing share and in final pick. It cuts
re-testing in the over-confirming models and lifts two of their picks by more than three units. It moves the re-testing of
Gemini 3.1 Pro and Sol by at most four points. \emph{Under-confirmation.} (d) The grid campaigns of Gemini 3.1 Pro and Sol, replayed with the last round spent re-testing the top $m$ candidates. With $m = 1$ the top one is re-tested four times, and with $m = 4$ each of the top four once. The y-axis is the change in the
true value of the final pick against the agent's own pick. Each dotted line is that model's selection loss, the most better checking could win back. Gemini 3.1 Pro
recovers about a quarter of it; Sol gains nothing.}
\label{fig:fixes}
\end{figure}

To understand why the three over-confirming models re-test so much (Figure~\ref{fig:headline}), we tested three
explanations on the grid. Each comes with an aid, given to these three models on the same seeds as their control, and
if an explanation holds, its aid should lower re-testing. GPT-6.1 Sol and Gemini 3.1 Pro do not over-confirm, so these
aids do not apply to them.
\begin{enumerate}
\item \textbf{Ledger.} \emph{They lose track of what they have measured.} The agent sees its history as one line per order, so a
candidate's readings are scattered across rounds. The ledger lists each measured candidate with its number of readings,
their mean and its standard error.
\item \textbf{Landscape note.} \emph{They do not know where to look next.} The agent is told nothing about how the landscape is built. The landscape note states how a candidate's score is built from its letters, and that a viable candidate's
neighbours are far more often viable than candidates in general.
\item \textbf{Menu.} \emph{They run out of new candidates worth trying.} The agent must name every new candidate itself, and most of
the new names it tries are rejected as non-viable. The Menu lists ten candidates the agent has never ordered, each one
letter from one of its three best, ranked by a simple model fitted to its own readings.
\end{enumerate}
The ledger and the landscape note changed re-testing by at most five percentage points (Figure~\ref{fig:fixes}a).
Three of these six changes were rises: the ledger for Opus 4.7 and DeepSeek V3.1, and the landscape note for DeepSeek V3.1. The Menu cut re-testing sharply in all three models, raised the best candidate they measured and worsened
their selection (Appendix~\ref{app:split}). The
over-confirmers re-test because they run short of candidates worth testing.

We then gave the Menu to all five models on scrambled GB1, built from real protein data, to test whether it carries
over and lifts the final pick. It cut re-testing again in the three over-confirmers (Figure~\ref{fig:fixes}b). Without it,
Gemini 2.5 Flash and DeepSeek V3.1 tested on average only \nGbDistinctFlash{} and \nGbDistinctDeep{} different candidates per
campaign. With it, their picks rose by more than three units of true value (Figure~\ref{fig:fixes}c). Almost all of the gain
came from finding better candidates, that is, from a smaller exploration loss.

For Gemini 3.1 Pro and Sol the Menu moved re-testing by at most four points and did not change their picks detectably
(Figure~\ref{fig:fixes}b and c). They do not lack candidates. Gemini 3.1 Pro loses in how it picks, on single readings
(Section~\ref{sec:under}).

\subsection{Under-confirmation}
\label{sec:under}

Gemini 3.1 Pro under-confirms: it re-tests too little, so it often picks a candidate it measured only once and loses at selection (Figure~\ref{fig:loss}). If that is the cause, re-testing its top candidates before it picks should win back part
of the loss. Under-confirmation shows only on the grid, where re-testing pays, so we ran this test there. We asked whether the agent's final pick would have been better had it spent its last round re-checking its best candidates. To answer this without calling the model again, we rebuilt each grid campaign of Gemini 3.1 Pro, and of Sol for comparison, from its log,
which reproduces every reading exactly. We then replaced the final round's four orders with re-tests of the top $m$
candidates by mean reading, under the same noise and failure rate. A fixed rule made the pick, the study's registered selection rule. It ranks
candidates by average reading, and pulls the average of a candidate with few readings toward the campaign's overall
average, so one lucky reading counts for less.

Gemini 3.1 Pro's pick improves as the re-tests are spread over more candidates, and at $m = 3$ it recovers about
\nAoneProRecoveredKthree\% of its selection loss (Figure~\ref{fig:fixes}d). This fits a true best that often sits just behind
a candidate with one lucky reading, so a second reading lets it overtake. The same rule applied to its unchanged readings gains almost nothing (Appendix~\ref{app:aone}), so the gain comes
from the extra readings. The gain stays small because Pro ends a campaign with
dozens of candidates measured only once, and four slots can re-check only a few. Sol gains nothing, since its top
candidates already have more than one reading each. Its curve dips slightly below zero because the replay gives up the new
candidates its real final round would have tried. In an exploratory run, re-testing over the last two rounds
recovered about as much as the last round alone.

\section{Wellocate as a benchmark and RL environment}
\label{sec:bench}

Our study measures how an agent splits a noisy budget between new candidates and re-tests. We release Wellocate so
others can measure their own agents the same way, and train agents to split the budget better. It runs our campaigns
through a standard reinforcement-learning interface and gives language models the exact prompts ours received. It comes with frozen splits of scrambled GB1 and PhoQ and of generated grid landscapes (Appendix~\ref{app:using}). The environment,
its code, the campaign logs and the analysis behind each number in this paper are at
\url{https://github.com/Xstendyl/wellocate}.

\begin{figure}[tbp]
\centering
\includegraphics[width=\textwidth]{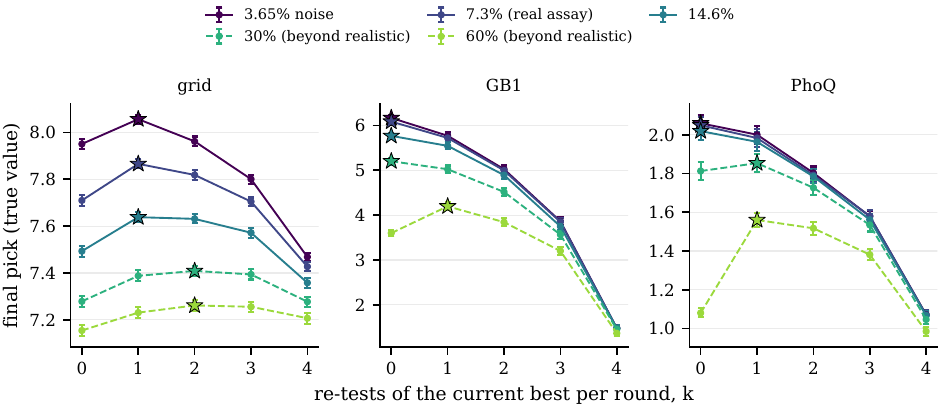}
\caption{The guessing rule's final pick for each number of re-tests per round, with 95\% intervals. Solid curves
cover realistic noise; dashed curves go far beyond it, to show where re-testing starts to pay on the proteins. The star
marks each curve's best setting; where a curve is flat at the top, the star is a near tie.}
\label{fig:noise}
\end{figure}

An environment for this problem is only worth learning in if the right split depends on the situation. If one fixed
rule were best everywhere, there would be nothing to learn. On each landscape,
at five noise levels, we ran the guessing rule at every setting from no re-tests to four a round. We recorded how good its final pick was (Figure~\ref{fig:noise}).

Each curve is one noise level, and its peak, marked with a star, is the best number of re-tests. At the real assay's
noise the grid peaks at one re-test a round. GB1 and PhoQ peak at none, and every re-test added makes the pick worse.
More noise moves the peak: re-testing begins to pay on PhoQ at about \nPhoqTimesLow{} times the assay's noise and on
GB1 only at about \nGbTimesHigh{} times. The reason is how close the best candidates sit. The ten best grid candidates
lie within \nTopSpreadGrid{} noise standard deviations of the best, against \nTopSpreadGb{} on GB1 and \nTopSpreadPq{} on PhoQ. One reading therefore rarely puts protein candidates in the wrong order.

The best fixed rule changes with the landscape and the noise. A good policy has to read the landscape and the noise and set its split from them,
and Wellocate asks a learner to do exactly that. This is also why Section~\ref{sec:res} judges each model against the
best rule for its own landscape and noise.

Last, we asked whether an agent can learn this task inside Wellocate. We trained a small policy from scratch with proximal policy optimisation (PPO), a standard reinforcement-learning method,
only on generated grid landscapes. It never saw a protein or a language model's campaign (Appendix~\ref{app:b11}).
On grid landscapes it had not seen, its final pick beats a policy that orders untried candidates at random by
\nPolRandGrid{} units of true value, and matches the guessing rule. Its lead is not clear on the most rugged generated landscapes. It also works on both proteins, under scrambled and
real names, and stays below the best scripted rule there. In a later check, given the same landscapes, seeds, noise and budget as
each language model's grid campaigns, it picks better than all five models. Replays show what it learned: it builds new
candidates from its best ones and almost never re-tests them. A small learned policy can already beat these models on the
grid, and learning when to confirm is still open.

\section{Discussion}
\label{sec:discussion}

Our results suggest that current language models do not split a noisy experimental budget well, and that they go wrong
in two directions. Three of the five models re-test far more than any scripted rule and run short of new candidates
worth trying. Gemini 3.1 Pro re-tests too little and often takes forward a candidate it measured once. Both failures
stay hidden in the final answer, which always looks like a reasonable pick. What matters is the useful output of every
slot over the whole campaign: an agent that re-tests everything rarely backs a lucky reading, and it also finds little.

The failures sit at the level of the campaign, and more thought about each experiment did not fix them. Gemini 3.1 Pro
sometimes reasoned for over 16{,}000 tokens before one round of orders, and its final pick still had a single reading in
\nProOneReading\% of grid campaigns. Opus 4.7 and Gemini 2.5 Flash still re-test most of
their slots with reasoning switched on (Appendix~\ref{app:panels}). Prompt wording did not fix them either. The one wording change that passed our registered test shrank to almost
nothing on fresh seeds (Appendix~\ref{app:replication}). A one-sentence lab role lowered re-testing by at most
ten points (Appendix~\ref{app:bio}). What helped the over-confirmers was a cheap supply of
candidates worth testing (Section~\ref{sec:why}). For the under-confirmer, one final round of checks recovered about a
quarter of its loss, which suggests confirmation has to run through the whole campaign (Section~\ref{sec:under}).

For a lab running agents, this suggests three steps.
\begin{enumerate}
\item \emph{Measure the split.} Compare the agent's re-testing share with the best scripted share for its landscape
and noise, and split its loss into exploration and selection. Wellocate's scorecard reports all three.
\item \emph{Give an over-confirmer candidates.} A cheap candidate generator, such as the Menu, lets it spend the freed
slots on new candidates.
\item \emph{Make an under-confirmer re-test along the way.} Re-test a leading candidate every round and the top
few at the end, as the scripted model-guided rule does.
\end{enumerate}

The main limits of our study are the range of landscapes and models we tested and the simple agent we used. Appendix~\ref{app:env} lists the parts of a real lab that Wellocate leaves out. We compared five models on three landscapes, one
generated and two real proteins under scrambled names, so other real landscapes may behave differently. Among the
proteins the Menu was tested on GB1 only, and under-confirmation appeared in one model, Gemini 3.1 Pro, on the grid
alone. Our agent was deliberately simple: each round it made one fresh call with the full history in its prompt, with no
tools and no other memory. An agent with tools, memory or a planning layer may split its budget differently.

\section{Related work}
\label{sec:related}

Language-model agents already plan and run experiments in chemistry \citep{boiko2023coscientist}. In biology they
design genetic perturbation screens, tested against data from screens already completed \citep{roohani2024biodiscovery}.
\citet{biodesignbench2026} test agents on protein design and trace their shortfall to how thoroughly they check
candidates. The agents pick suitable tools, score each candidate on only a few measures and stop exploring early. In
those tasks chance enters through design tools that sample new candidates. In ours it enters through measurement, so
one candidate measured twice gives two different numbers. \citet{wang2026agentnoise} add noise to what users say and
what tools return in existing agent benchmarks, and measure how well agents hold up.

\citet{krishnamurthy2024explore} placed GPT-3.5, GPT-4 and Llama2 in multi-armed bandits, where an agent repeatedly
picks one of a fixed set of options with random payoffs. The prompt held the task and the history of past picks, either raw
or summarised. Only GPT-4, reasoning step by step with a summarised history, explored well. Our ledger gives the same
kind of summary, and in our study it changed re-testing little (Section~\ref{sec:why}). Their payoffs are coin flips,
so the noise level is tied to the payoff and cannot be set on its own.

LLAMBO puts a language model inside Bayesian optimisation, to propose candidates and predict their scores
\citep{liu2024llambo}. \citet{gupta2025bayesopt} tested whether such a model uses the results it gets: shuffling the
measured outcomes on genetic perturbation tasks left performance essentially unchanged. On feedback from a real CRISPR
screen, \citet{wainrib2026labloop} found that shuffling removed the gain. In both studies the results come from
completed screens, and re-measuring a gene is not offered as a use of the budget.

Experimental design has long studied how to split a budget between new conditions and repeats.
\citet{binois2019replication} study when repeating a run beats trying a new one in noisy computer simulations.
\citet{chen2000ocba} split a fixed number of simulation runs among a set of options so as to pick the best one.
\citet{dai2023replicable} describe the trade-off our agents face, under large noise whose size varies across
conditions. One can test more new conditions with fewer repeats each, or fewer conditions with more repeats. Their
method sets the number of repeats for each chosen condition from the data so far. Our scripted rules are simpler: each
re-tests its current best a fixed number of times per round. A rule that sets its re-testing from the data would be a
stronger yardstick. In bandit terms, a Wellocate campaign is close to best-arm identification on a fixed budget. The
task is to find the best option with a set number of noisy tries, here made four at a time. The score is the true value of the final pick. In Wellocate, candidates that share letters have related values, so one
reading says something about other candidates, and the agent may nominate a candidate it never measured.

\citet{wu2016adaptation} measured the fitness of nearly every variant at four sites of the protein GB1 and mapped its
fitness peaks. Our GB1 track is built from their data. \citet{wu2019mlde} used the same landscape to test
machine-learning-guided directed evolution, which picks the next variants to measure from a model fitted to earlier
readings. Our model-guided rule is a simple version of that idea. Our grid uses the NK model \citep{kauffman1989nk}, a
classic generated landscape. Other generated landscapes come from physical models: \citet{lobkovsky2011predictability}
build them from a protein-folding model and ask how predictable evolution is on them.

We find four benchmark environments closest to our work. Wellocate combines four features: landscapes generated by program,
measurement noise whose level we set, down to zero, re-testing as a choice that uses budget, and one scored final pick.
BoxingGym \citep{gandhi2025boxinggym} is the nearest. Its ten environments are probabilistic models of fixed real-world
domains, and the agent's goal is to learn the model. Experiments are scored by expected information gain, how much
each is expected to teach about the model. The study does not vary the noise level or test a noise-free setting, and
it does not study repeating an experiment as a choice. \citet{park2026exploration} also generate their environments by
program and aim to isolate specific behaviours of agents. Their observations carry no noise, so a second look returns
what the first one showed. \citet{wang2026allocbench} have agents allocate a fixed budget, under task descriptions that
range from abstract to concrete, and their outcomes carry no noise. \citet{huai2026agenthpo} study agents that decide
from accumulated experimental results, on 30 real machine-learning tasks with no set noise level. Each of these
lacks at least one of the four features, and each keeps its noise level fixed.

\section*{Acknowledgements}

We used AI coding assistants for implementation, analysis and drafting. The authors checked all results and take full responsibility for the content.

\bibliography{refs}

\begin{thebibliography}{22}
\providecommand{\natexlab}[1]{#1}
\providecommand{\url}[1]{\texttt{#1}}
\expandafter\ifx\csname urlstyle\endcsname\relax
  \providecommand{\doi}[1]{doi: #1}\else
  \providecommand{\doi}{doi: \begingroup \urlstyle{rm}\Url}\fi

\bibitem[Binois et~al.(2019)Binois, Huang, Gramacy, and
  Ludkovski]{binois2019replication}
Mickael Binois, Jiangeng Huang, Robert~B. Gramacy, and Mike Ludkovski.
\newblock Replication or exploration? {S}equential design for stochastic
  simulation experiments.
\newblock \emph{Technometrics}, 61\penalty0 (1):\penalty0 7--23, 2019.
\newblock \doi{10.1080/00401706.2018.1469433}.
\newblock arXiv:1710.03206.

\bibitem[Boiko et~al.(2023)Boiko, MacKnight, Kline, and
  Gomes]{boiko2023coscientist}
Daniil~A. Boiko, Robert MacKnight, Ben Kline, and Gabe Gomes.
\newblock Autonomous chemical research with large language models.
\newblock \emph{Nature}, 624\penalty0 (7992):\penalty0 570--578, 2023.
\newblock \doi{10.1038/s41586-023-06792-0}.

\bibitem[Chen et~al.(2000)Chen, Lin, Yücesan, and Chick]{chen2000ocba}
Chun-Hung Chen, Jianwu Lin, Enver Yücesan, and Stephen~E. Chick.
\newblock Simulation budget allocation for further enhancing the efficiency of
  ordinal optimization.
\newblock \emph{Discrete Event Dynamic Systems}, 10\penalty0 (3):\penalty0
  251--270, 2000.
\newblock \doi{10.1023/A:1008349927281}.

\bibitem[Dai et~al.(2023)Dai, Nguyen, Tay, Urano, Leong, Low, and
  Jaillet]{dai2023replicable}
Zhongxiang Dai, Quoc~Phong Nguyen, Sebastian~Shenghong Tay, Daisuke Urano,
  Richalynn Leong, Bryan Kian~Hsiang Low, and Patrick Jaillet.
\newblock Batch bayesian optimization for replicable experimental design.
\newblock In \emph{Advances in Neural Information Processing Systems},
  volume~36, 2023.

\bibitem[Gandhi et~al.(2025)Gandhi, Li, Goodyear, Bhatia, Li, Bhaskar, Zaman,
  and Goodman]{gandhi2025boxinggym}
Kanishk Gandhi, Michael~Y. Li, Lyle Goodyear, Agam Bhatia, Louise Li, Aditi
  Bhaskar, Mohammed Zaman, and Noah~D. Goodman.
\newblock {BoxingGym}: Benchmarking progress in automated experimental design
  and model discovery.
\newblock \emph{arXiv preprint arXiv:2501.01540}, 2025.

\bibitem[Gupta et~al.(2025)Gupta, Hartford, and Liu]{gupta2025bayesopt}
Rushil Gupta, Jason Hartford, and Bang Liu.
\newblock {LLMs} for bayesian optimization in scientific domains: Are we there
  yet?
\newblock In \emph{Findings of the Association for Computational Linguistics:
  EMNLP 2025}, pages 15482--15510, 2025.
\newblock \doi{10.18653/v1/2025.findings-emnlp.838}.

\bibitem[Huai et~al.(2026)Huai, Fan, Chen, Zheng, Wang, Chen, Zhou, and
  Huang]{huai2026agenthpo}
Tianyu Huai, Tingshuo Fan, Xinchi Chen, Yining Zheng, Yuxin Wang, Shuang Chen,
  Jie Zhou, and Xuanjing Huang.
\newblock {AgentHPOBench}: A benchmark for evaluating {LLM} agents as
  sequential hyperparameter optimizers.
\newblock \emph{arXiv preprint arXiv:2607.29626}, 2026.

\bibitem[Kauffman and Weinberger(1989)]{kauffman1989nk}
Stuart~A. Kauffman and Edward~D. Weinberger.
\newblock The {NK} model of rugged fitness landscapes and its application to
  maturation of the immune response.
\newblock \emph{Journal of Theoretical Biology}, 141\penalty0 (2):\penalty0
  211--245, 1989.
\newblock \doi{10.1016/S0022-5193(89)80019-0}.

\bibitem[Kim and Romero(2026)]{biodesignbench2026}
Jeonghyeon Kim and Philip Romero.
\newblock Benchmarking and behavioral characterization of {LLM} agents for
  protein design.
\newblock \emph{bioRxiv}, 2026.
\newblock \doi{10.64898/2026.05.06.723381}.

\bibitem[Krishnamurthy et~al.(2024)Krishnamurthy, Harris, Foster, Zhang, and
  Slivkins]{krishnamurthy2024explore}
Akshay Krishnamurthy, Keegan Harris, Dylan~J. Foster, Cyril Zhang, and
  Aleksandrs Slivkins.
\newblock Can large language models explore in-context?
\newblock In \emph{Advances in Neural Information Processing Systems},
  volume~37, pages 120124--120158, 2024.
\newblock \doi{10.52202/079017-3818}.

\bibitem[Liu et~al.(2024)Liu, Astorga, Seedat, and van~der
  Schaar]{liu2024llambo}
Tennison Liu, Nicolás Astorga, Nabeel Seedat, and Mihaela van~der Schaar.
\newblock Large language models to enhance bayesian optimization.
\newblock In \emph{The Twelfth International Conference on Learning
  Representations (ICLR)}, 2024.
\newblock URL \url{https://openreview.net/forum?id=OOxotBmGol}.

\bibitem[Lobkovsky et~al.(2011)Lobkovsky, Wolf, and
  Koonin]{lobkovsky2011predictability}
Alexander~E. Lobkovsky, Yuri~I. Wolf, and Eugene~V. Koonin.
\newblock Predictability of evolutionary trajectories in fitness landscapes.
\newblock \emph{PLoS Computational Biology}, 7\penalty0 (12):\penalty0
  e1002302, 2011.
\newblock \doi{10.1371/journal.pcbi.1002302}.

\bibitem[Park et~al.(2026)Park, Kim, Jeong, Nowak, Lee, and
  Lee]{park2026exploration}
Jaden Park, Jungtaek Kim, Jongwon Jeong, Robert~D. Nowak, Kangwook Lee, and
  Yong~Jae Lee.
\newblock Exploration and exploitation errors are measurable for language model
  agents.
\newblock \emph{arXiv preprint arXiv:2604.13151}, 2026.

\bibitem[Podgornaia and Laub(2015)]{podgornaia2015phoq}
Anna~I. Podgornaia and Michael~T. Laub.
\newblock Pervasive degeneracy and epistasis in a protein-protein interface.
\newblock \emph{Science}, 347\penalty0 (6222):\penalty0 673--677, 2015.
\newblock \doi{10.1126/science.1257360}.

\bibitem[Rapp et~al.(2024)Rapp, Bremer, and Romero]{rapp2024selfdriving}
Jacob~T. Rapp, Bennett~J. Bremer, and Philip~A. Romero.
\newblock Self-driving laboratories to autonomously navigate the protein
  fitness landscape.
\newblock \emph{Nature Chemical Engineering}, 1\penalty0 (1):\penalty0 97--107,
  2024.
\newblock \doi{10.1038/s44286-023-00002-4}.

\bibitem[Roohani et~al.(2025)Roohani, Lee, Huang, Vora, Steinhart, Huang,
  Marson, Liang, and Leskovec]{roohani2024biodiscovery}
Yusuf Roohani, Andrew Lee, Qian Huang, Jian Vora, Zachary Steinhart, Kexin
  Huang, Alexander Marson, Percy Liang, and Jure Leskovec.
\newblock {BioDiscoveryAgent}: An {AI} agent for designing genetic perturbation
  experiments.
\newblock In \emph{International Conference on Learning Representations
  (ICLR)}, 2025.
\newblock URL \url{https://openreview.net/forum?id=HAwZGLcye3}.

\bibitem[Tom et~al.(2024)Tom, Schmid, Baird, Cao, Darvish, Hao, Lo,
  Pablo-García, Rajaonson, Skreta, Yoshikawa, Corapi, Akkoc, Strieth-Kalthoff,
  Seifrid, and Aspuru-Guzik]{tom2024selfdriving}
Gary Tom, Stefan~P. Schmid, Sterling~G. Baird, Yang Cao, Kourosh Darvish, Han
  Hao, Stanley Lo, Sergio Pablo-García, Ella~M. Rajaonson, Marta Skreta,
  Naruki Yoshikawa, Samantha Corapi, Gun~Deniz Akkoc, Felix Strieth-Kalthoff,
  Martin Seifrid, and Alán Aspuru-Guzik.
\newblock Self-driving laboratories for chemistry and materials science.
\newblock \emph{Chemical Reviews}, 124\penalty0 (16):\penalty0 9633--9732,
  2024.
\newblock \doi{10.1021/acs.chemrev.4c00055}.

\bibitem[Wainrib et~al.(2026)Wainrib, Bodinier, Dakhli, Monserrat, Perez,
  Carpentier, Codato, and Klein]{wainrib2026labloop}
Gilles Wainrib, Barbara Bodinier, Haitem Dakhli, Josep Monserrat,
  Almudena~Espin Perez, Sabrina Carpentier, Roberta Codato, and John Klein.
\newblock Can {AI} scientist agents learn from lab-in-the-loop feedback?
  evidence from iterative perturbation discovery.
\newblock \emph{arXiv preprint arXiv:2603.26177}, 2026.

\bibitem[Wang and Xu(2026)]{wang2026allocbench}
Daniel Wang and Andrew Xu.
\newblock {AllocBench}: Measuring online tool allocation capability in {LLM}
  agents.
\newblock \emph{arXiv preprint arXiv:2607.23332}, 2026.

\bibitem[Wang et~al.(2026)Wang, Chen, Wang, Wu, Fang, Cai, Gu, Su, Zhang, Wang,
  Cai, and Chua]{wang2026agentnoise}
Ruipeng Wang, Yuxin Chen, Yukai Wang, Chang Wu, Junfeng Fang, Xiaodong Cai,
  Qi~Gu, Hui Su, An~Zhang, Xiang Wang, Xunliang Cai, and Tat-Seng Chua.
\newblock {AgentNoiseBench}: Benchmarking robustness of tool-using {LLM} agents
  under noisy condition.
\newblock \emph{arXiv preprint arXiv:2602.11348}, 2026.

\bibitem[Wu et~al.(2016)Wu, Dai, Olson, Lloyd-Smith, and Sun]{wu2016adaptation}
Nicholas~C Wu, Lei Dai, C~Anders Olson, James~O Lloyd-Smith, and Ren Sun.
\newblock Adaptation in protein fitness landscapes is facilitated by indirect
  paths.
\newblock \emph{eLife}, 5:\penalty0 e16965, 2016.
\newblock \doi{10.7554/eLife.16965}.

\bibitem[Wu et~al.(2019)Wu, Kan, Lewis, Wittmann, and Arnold]{wu2019mlde}
Zachary Wu, S.~B.~Jennifer Kan, Russell~D. Lewis, Bruce~J. Wittmann, and
  Frances~H. Arnold.
\newblock Machine learning-assisted directed protein evolution with
  combinatorial libraries.
\newblock \emph{Proceedings of the National Academy of Sciences}, 116\penalty0
  (18):\penalty0 8852--8858, 2019.
\newblock \doi{10.1073/pnas.1901979116}.

\end{thebibliography}

\clearpage
\appendix
\section*{Appendices}
\providecommand{\estci}[2]{\begin{tabular}[t]{@{}c@{}}#1\\[-1pt]{\scriptsize #2}\end{tabular}}
The appendices fall into four groups. Appendices~\ref{app:env} to \ref{app:splits} describe the environment, the
scripted rules and how to use Wellocate. Appendices~\ref{app:amendments} to \ref{app:aone} hold the five-model study: what was registered, the runs, PhoQ and the checks behind Section~\ref{sec:res}.
Appendices~\ref{app:grid} to \ref{app:sim} keep the pre-registered prompt study, its follow-up experiments and the learnability simulation. Appendix~\ref{app:b11} covers the trained policy.

\section{Environment details}
\label{app:env}

\paragraph{The landscape.} A candidate's fitness comes from an NK landscape \citep{kauffman1989nk}. $N$ is
the number of positions, here four, and $K$ is how many other positions each position's contribution
depends on. A landscape is rugged when it has local peaks, candidates that no single-letter change can
improve, and a larger $K$ makes it more rugged. We set $K=1$, so each position's contribution depends on the
next position along the string, wrapping round at the end. Each position has its own table of
contributions, one for every pair of letters at that position and the next. The entries are drawn uniformly between 0 and 1 when the landscape is generated. A candidate's fitness is the mean of its four contributions. The whole
landscape is then multiplied by one constant so its maximum is 8.762, the highest value measured in the
published data. This only sets the units, and beyond that value and its share of working candidates the
landscape is a standard NK landscape. At $K=1$ position interactions are weak, so the search itself is simple, and an
agent that falls short does so through how it spends measurements. The agents were run at $K=1$ only, and
the learnability simulation in Appendix~\ref{app:sim} repeats the scripted analysis at $K=2$ and 3.

\paragraph{Viability.} A candidate is viable if its fitness exceeds the 93rd percentile, which leaves
11{,}199 viable candidates, 7.00\% of the space. The threshold was chosen so that about the same share of
candidates works as in the published data. In GB1, \nGbViable\% of variants keep at least a tenth of the wild type's
fitness. GB1 and PhoQ are cut at the same share, so their never-measured variants, stored as zero, are all
non-viable. On PhoQ the authors' own call of functional variants is much narrower, so there the cut also admits
variants they would call non-functional. Treating ``does not work'' as a sharp cutoff on fitness is a
simplification.

\paragraph{Ordering and failures.} Each slot is filled by an order, a request to measure one candidate
once. An order for a non-viable candidate is rejected. The agent is told, and is asked for a replacement in
the same round, up to three times. If every replacement is also non-viable, the slot is lost and still
counts against the budget. About 15\% of orders fail at random, whatever the candidate's fitness, and a
failed order returns nothing and still uses its slot. The rate is a design choice. Over the 5{,}075 agent
campaigns of the prompt study (Appendix~\ref{app:grid}), 14.97\% of 300{,}898 measurements attempted on viable candidates failed.

\paragraph{Noise.}
\label{app:noise}
Each measurement of a viable candidate returns its true fitness plus independent Gaussian error, whose
standard deviation grows with the true fitness $f$:
\[ \sigma(f) = \sqrt{0.00351^2 + (0.073 f)^2}. \]
Both constants in $\sigma(f)$ are sized from the read counts of \citet{wu2016adaptation}. They screened a library
spanning all 160{,}000 variants in one deep mutational scan and got a usable reading for 93.4\% of them. In such an experiment every variant is mixed into one
pool, and the pool goes through a selection step that keeps variants binding a target. A variant that binds
better is kept in larger numbers. Each protein stays physically linked to the DNA that encodes it, so
counting that DNA counts the protein. The DNA is counted before and after selection by sequencing, which
reads DNA in short pieces. Each piece is one read, a variant's count is how many reads matched it, and the
reads per variant is the \emph{read depth}.

In that study a variant's fitness is its after-to-before read ratio, divided by the same ratio for the
unmodified protein the study started from. It has nothing to do with \texttt{GDIF}, the grid's reference candidate. The study's
published read totals average 557 reads per variant before selection and 285 after. Treating each count as Poisson, the variance
of the log ratio is about $1/285 + 1/557$, so its standard deviation, the relative error, is about 7.3\%. We apply this
average depth to every candidate, which gives the constant 0.073. Real counts are often more spread out than Poisson, so this is a
lower estimate of that study's counting error.

Taking the starting protein's ratio as the study's average, $285/557$, a variant with 557 reads before
selection and a single read after has fitness $(1/557)/(285/557)$, or 0.0035 units. That is the smallest
non-zero value the study could resolve, and we use it as the smallest spread any reading has. It matters
only near zero fitness, so it has no practical effect on a viable candidate.

A variant that binds better gets more reads, so one average depth overstates the counting error of the best
candidates. A real re-run also varies for reasons beyond counting. The simulation may therefore be noisier or less
noisy than a real lab, and which of the two holds is not known.

\paragraph{The reference candidate.} \texttt{GDIF} is the viable candidate at the median fitness. Its true
fitness, 6.8730, is stated to the agent to two decimals, and Table~\ref{tab:cells} gives the re-testing
share with and without re-orders of it. We also checked that the scripted results do not depend on this
start. On each landscape we drew \nStartsRandom{} viable starts at random and ran the three scripted rules from each,
on our campaign seeds. The guessing and informed rules keep \texttt{GDIF}'s best number of re-tests from every start on all three landscapes. On the grid the guessing rule's re-testing share at that setting stays
between \nStartGuessShareMinGrid\% and \nStartGuessShareMaxGrid\%. On the grid the model-guided rule's best setting moves
between no re-test and one a round, which score almost the same. At our setting of one re-test its share stays between \nStartMgShareMinGrid\% and \nStartMgShareMaxGrid\%. Keeping that setting costs a median of
\nStartMgPickLostMedGrid{} in final pick, at most \nStartMgPickLostMaxGrid. On GB1 and PhoQ it keeps \texttt{GDIF}'s setting from \nStartSameKGbMg{} and \nStartSameKPqMg{} of the \nStartsRandom{} starts.

\paragraph{What the simulation leaves out.} A non-viable order is caught and replaced. In a lab, finding out would
cost a well. A new variant and a re-test cost the same slot. A lab must first make a new variant, which makes exploring dearer there and could raise a lab's best rate of re-testing. A lab
usually wants a shortlist, and ranking a shortlist could reward more re-testing than one final pick does.
Real rounds use plates of 96 or 384 wells, so the best number of repeats found here does not transfer as a
number. Whether the agents' misallocation depends on batch size or budget is not tested. Noise here has no plate effects
or drift, meaning slow shifts in readings between plates or over time, and in a real lab a repeat can also
catch those. So an agent that re-tests more than the scripted rules here would not necessarily re-test too much in a
real lab.

\section{The scripted rules}
\label{app:refs}

The guessing and informed rules of Section~\ref{sec:env} were the prompt study's references, and the model-guided
rule was added later under a registered change. The guessing and informed rules spend $k$ of the four slots a round
re-testing their current best, and $k^\ast$ is the best setting. The informed rule is told which candidates are
viable. It is never rejected, so it loses no slots to rejections, and it still has to find out how good each candidate is. The agents are not told
which candidates are viable, so the informed rule has an easier task than theirs. The guessing rule was pre-registered. It tries every candidate one letter away from its current best, and follows the agents' rules for rejections,
replacements and lost slots, in the same code. Both ran on the same 1{,}500 seeds per setting.

The model-guided rule spends its first four rounds trying one-letter changes of the start, one at each position. From
the fifth round it orders three new candidates a round, ranked by its letter scores. It also re-tests one leading candidate: its best, or the next best once the best has two readings. In its last three rounds it re-tests its four
best candidates by mean reading. It picks with the registered selection rule of Section~\ref{sec:under}.

\begin{table}[h]
\centering
\small
\caption{The guessing and informed rules on the grid. One knows more than the agents, the other knows the same.}
\begin{tabular}{lrr}
\toprule
 & informed (told which are viable) & guessing (must guess) \\
\midrule
best re-tests a round, $k^\ast$     & 1 of 4 slots & 1 of 4 slots \\
mean pick at $k{=}0$                & 7.8862 & 7.7098 \\
mean pick at $k^\ast$               & 7.9993 & 7.8659 \\
distinct candidates at $k^\ast$     & 39.14  & 28.94 \\
rejections per campaign at $k{=}1$  & 0.00   & 83.02 \\
lost slots per campaign at $k{=}1$  & 0.00   & 11.70 \\
campaigns collapsing at $k{=}4$     & 16 of 1{,}500 & 147 of 1{,}500 \\
\bottomrule
\end{tabular}
\end{table}

On the grid both do best at one re-test a round. Every comparison between the agents and the scripted rules there reaches the same conclusion with either rule.

\section{Using Wellocate}
\label{app:using}
\label{app:splits}

\paragraph{Interface.} An episode is one campaign. A call to \texttt{step} orders up to four candidates, or one
replacement after a rejection, and \texttt{nominate} ends the episode. Observations hold readings, outcomes and the
remaining budget, and no true value; the reward is the nominee's true value. A text mode returns the exact prompts our agents received and reads replies the same way, asking again when a reply cannot be read. A language model and a policy trained by reinforcement learning thus face the same task. The environment runs the same code as our experiments. Through it the
guessing rule reproduces our reference pick of 7.8659 and re-testing share of 23.78\% exactly, and the informed rule its pick of 7.9993. One core runs about \nEpisodesPerSec{} episodes a second.

\paragraph{Tracks and splits.} The protein track uses scrambled GB1 and PhoQ, with fresh scrambles for training,
validation and testing and the real amino-acid names held out (Table~\ref{tab:splits}). A scramble changes only the names, so a protein test split asks for unseen names of the same protein. Training on one protein and testing on the other asks for transfer. The grid track gives fresh NK landscapes and holds out rougher ones. Every split is fixed and public. A hidden test server is future work.

\paragraph{Scoring and baselines.} The scorecard gives each policy's mean pick and its regret against the informed rule on the same episodes. It adds the re-testing share, the exploration and selection losses, the candidates measured and the slots lost, each with a 95\% interval. Beside them it names the best scripted share for that landscape and noise.
The baselines are random ordering, the guessing rule, the informed rule and the model-guided rule. A fifth, a simple untuned confirmation policy, explores without repeats and re-tests its top candidates at the end. The scorecard also lists our language-model results for comparison.

\begin{figure}[htbp]
\centering
\includegraphics[width=\textwidth]{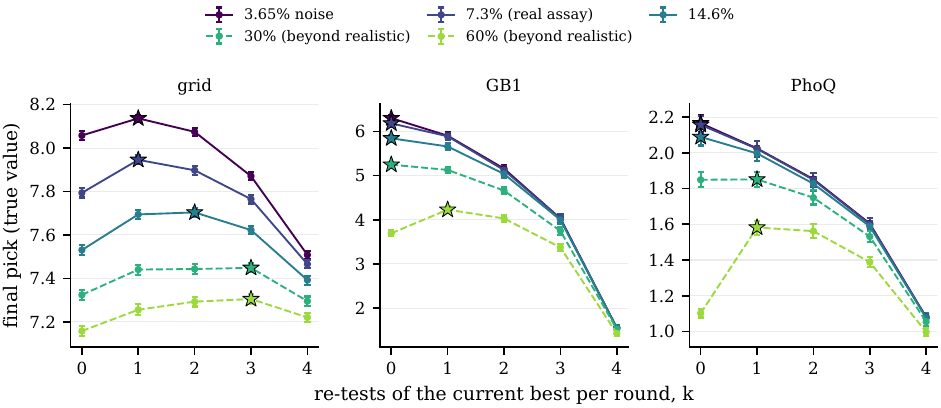}
\caption{Figure~\ref{fig:noise} with free rejections: a rejected order never costs a slot.}
\label{fig:noisefree}
\end{figure}

\paragraph{Noise and the best number of re-tests.} On the grid the guessing rule that never re-tests loses \nGridSelLow{},
\nGridSelAssay{} and \nGridSelHigh{} at selection at half, one and two times assay noise. On PhoQ one re-test a round ties with
no re-tests at about \nPhoqTimesLow{} times assay noise, and wins clearly at \nPhoqTimesHigh{} times. On GB1 it wins
only at \nGbTimesHigh{} times (Table~\ref{tab:onevszero}). Part of re-testing's value on the grid is avoiding
rejections: the rule that never re-tests loses \nGridLostKzero{} slots to rejections, against \nGridLostKone{} for one
re-test a round. With free rejections, one re-test a round still beats no re-tests on the grid at assay noise
(Table~\ref{tab:onevszero}), so there re-testing also pays because of the noise. With free rejections the best number
of re-tests also rises at lower noise on the grid, and GB1 and PhoQ keep the best settings of
Figure~\ref{fig:noise} (Figure~\ref{fig:noisefree}).

\begin{table}[tbp]
\centering
\small
\caption{The guessing rule's final pick with no re-tests and with one re-test a round, with 95\% intervals.}
\label{tab:onevszero}
\begin{tabular}{lcc}
\toprule
 & no re-tests & one re-test a round \\
\midrule
GB1, \nGbTimesHigh{} times assay noise & \estci{\nGbHighKzero}{\nGbHighKzeroci} & \estci{\nGbHighKone}{\nGbHighKoneci} \\
grid, free rejections, assay noise & \estci{\nGridRfKzero}{\nGridRfKzeroci} & \estci{\nGridRfKone}{\nGridRfKoneci} \\
\bottomrule
\end{tabular}
\end{table}

\begin{table}[!htbp]
\centering
\small
\setlength{\tabcolsep}{5pt}
\caption{Wellocate's frozen splits: landscapes / episodes per split. Train, validation and test use disjoint landscape seeds (scrambles for GB1 and PhoQ, NK seeds for the grid) and disjoint campaign seeds. Held out: the real amino-acid names for GB1 and PhoQ, and two rougher NK settings, $K = 2$ and $K = 3$. \emph{Paper} is the landscape and seeds used in this study.}
\label{tab:splits}
\begin{tabular}{llrrrrr}
\toprule
track & landscape & train & validation & test & held out & paper \\
\midrule
protein & scrambled GB1 & 400 / 100{,}000 & 10 / 1{,}500 & 20 / 1{,}500 & 1 / 1{,}500 & 1 / 1{,}500 \\
protein & scrambled PhoQ & 400 / 100{,}000 & 10 / 1{,}500 & 20 / 1{,}500 & 1 / 1{,}500 & 1 / 1{,}500 \\
grid & generated NK & 400 / 100{,}000 & 10 / 1{,}500 & 20 / 1{,}500 & \begin{tabular}[t]{@{}r@{}}$K = 2$: 20 / 1{,}500 \\ $K = 3$: 20 / 1{,}500\end{tabular} & 1 / 1{,}500 \\
\bottomrule
\end{tabular}
\end{table}

\section{What was registered}
\label{app:amendments}

The prompt study's design, outcome, confirmatory contrasts and analysis were registered before it ran
(Appendix~\ref{app:grid}). Every later change was registered, with its date, before the data it governs. These changes added the aids experiment on the grid, with labels for a large, a small or no change in re-testing fixed in advance. They also added the landscape note, the model-guided rule and the Menu on scrambled GB1. Later changes added GPT-6.1 Sol and Gemini 3.1 Pro on the grid and on GB1, and the PhoQ runs for all five models with a named test. The last added the replication of the Decision-rule result, the learnability simulation and the test of the trained policy. A cap on repeat orders was also registered and then withdrawn before its main run. In trial runs the agents met the
cap by proposing new names, most of them non-viable. No seed was dropped from the GPT-6.1 Sol, Gemini 3.1 Pro or PhoQ runs. Exclusions in other runs are reported where they occur.
Every other analysis in these appendices was decided after the registered ones. It describes, checks or explores
the results and tests no claim, and so does the noise sweep of Section~\ref{sec:bench}.

\section{The five models}
\label{app:comparators}
All five models ran at temperature 0 (Table~\ref{tab:models}). OpenRouter sends each call for an open-weight
model to one of its hosting companies. In every run DeepSeek V3.1's calls went to five or six hosts, and its
results average over them. Opus 4.7, Gemini 2.5 Flash and DeepSeek V3.1 ran with
reasoning off. GPT-6.1 Sol and Gemini 3.1 Pro ran at their lowest reasoning setting. Even so, Gemini 3.1 Pro used more than 16{,}000 reasoning tokens on each of \nProHeavyCalls{} calls, up to \nProMaxReasoning{} on one. All five saw the Blind prompt of Appendix~\ref{app:grid}, which says nothing about measurement error or repeats.

\begin{table}[tbp]
\centering
\small
\caption{The five models as run on the grid. GPT-6.1 Sol and Gemini 3.1 Pro ran at their lowest reasoning setting. Reasoning tokens are the median over every logged call of the runs with the Blind prompt. Sol reported zero reasoning tokens on at least half of its calls.}
\label{tab:models}
\begin{tabular}{llll}
\toprule
model & released & reasoning as run & median reasoning tokens per call \\
\midrule
Opus 4.7 & 2026-02 & off & 0 \\
Gemini 2.5 Flash & 2025-06 & off & 0 \\
DeepSeek V3.1 & 2025-08 & off & 0 \\
GPT-6.1 Sol & 2026-09 & minimal & 0 \\
Gemini 3.1 Pro & 2026-02 & minimal & 240 \\
\bottomrule
\end{tabular}
\end{table}

\subsection{Results on the grid}

Table~\ref{tab:fivegrid} gives each model's re-testing share, final pick and losses.
The scripted band runs from \nGuessShare\% to \nBandHi\% of slots, or \nBandNonlostLo\% to \nBandNonlostHi\% counted over slots not lost. The guessing rule's exploration and selection losses are 0.67 and 0.23. Every model's pick
interval lies below the guessing rule's 7.87. Gemini 3.1 Pro's final pick had a single reading in
\nGridOneReadingPro\% of its campaigns, against \nGridOneReadingInf\% for the informed rule and
\nGridOneReadingGuess\% for the guessing rule. With reasoning on (Appendix~\ref{app:reasoning}), Opus 4.7 re-tests
\nPanelcOpus\% of its slots and Gemini 2.5 Flash \nPanelcFlash\%. The aids' effects on re-testing are in
Table~\ref{tab:fixes}.

\begin{table}[tbp]
\centering
\small
\caption{The five models on the grid, with 95\% intervals. The second column counts re-testing over slots not lost.}
\label{tab:fivegrid}
\begin{tabular}{lccccc}
\toprule
 & share (\%) & share, slots not lost (\%) & final pick & exploration loss & selection loss \\
\midrule
Opus 4.7 & \estci{\nShareOpus}{\nShareOpusci} & \estci{\nNonlostOpus}{\nNonlostOpusci} & \estci{\nPickOpus}{\nPickOpusci} & \estci{\nExplOpus}{\nExplOpusci} & \estci{\nSelOpus}{\nSelOpusci} \\
Gemini 2.5 Flash & \estci{\nShareFlash}{\nShareFlashci} & \estci{\nNonlostFlash}{\nNonlostFlashci} & \estci{\nPickFlash}{\nPickFlashci} & \estci{\nExplFlash}{\nExplFlashci} & \estci{\nSelFlash}{\nSelFlashci} \\
DeepSeek V3.1 & \estci{\nShareDeep}{\nShareDeepci} & \estci{\nNonlostDeep}{\nNonlostDeepci} & \estci{\nPickDeep}{\nPickDeepci} & \estci{\nExplDeep}{\nExplDeepci} & \estci{\nSelDeep}{\nSelDeepci} \\
GPT-6.1 Sol & \estci{\nShareSol}{\nShareSolci} & \estci{\nNonlostSol}{\nNonlostSolci} & \estci{\nPickSol}{\nPickSolci} & \estci{\nExplSol}{\nExplSolci} & \estci{\nSelSol}{\nSelSolci} \\
Gemini 3.1 Pro & \estci{\nSharePro}{\nShareProci} & \estci{\nNonlostPro}{\nNonlostProci} & \estci{\nPickPro}{\nPickProci} & \estci{\nExplPro}{\nExplProci} & \estci{\nSelPro}{\nSelProci} \\
\bottomrule
\end{tabular}
\end{table}

\begin{table}[!htbp]
\centering
\footnotesize
\caption{Change in re-testing share on the grid when each aid is added, in percentage points against the same model's control on the same seeds, with 95\% intervals. Negative values mean less re-testing. The Menu's effect on scrambled GB1 is in Figure~\ref{fig:fixes}.}
\label{tab:fixes}
\begin{tabular}{lrrr}
\toprule
model & ledger & landscape note & Menu \\
\midrule
Opus 4.7 & $+$4.4 [$+$2.6, $+$6.2] & $-$4.7 [$-$7.8, $-$1.5] & $-$18.2 [$-$20.8, $-$15.5] \\
Gemini 2.5 Flash & $-$0.9 [$-$2.1, $+$0.2] & $-$0.5 [$-$1.8, $+$0.8] & $-$26.2 [$-$27.9, $-$24.5] \\
DeepSeek V3.1 & $+$0.8 [$+$0.1, $+$1.6] & $+$1.5 [$+$0.8, $+$2.2] & $-$31.4 [$-$33.5, $-$29.4] \\
\bottomrule
\end{tabular}
\end{table}

Each model's numbers come from the runs that its registered comparison with the scripted rules uses. Gemini 2.5 Flash and DeepSeek V3.1 come from the control runs of the aids experiment, whose prompt options match those of the GPT-6.1 Sol and Gemini 3.1 Pro runs. Opus 4.7 comes from a rerun of the original prompts made on the same day as prompt check A (Appendix~\ref{app:prompt}). That rerun differs only in the message sent back when a reply cannot be read. Counting re-testing over slots not lost leaves the conclusions unchanged. The GPT-6.1 Sol and Gemini 3.1 Pro runs cost \$\nSolSpend{} and \$\nGemSpend, and the PhoQ runs \$\nPqSpend. The scripted rules and Wellocate cost nothing to rerun.

\subsection{Results on scrambled GB1}

These runs used different landscapes and seeds from the grid runs, so they do not pair with them; Table~\ref{tab:landscapes} gives their shares. Gemini 2.5 Flash measures \nGbDistinctFlash{}
distinct candidates and DeepSeek V3.1 \nGbDistinctDeep, both in the control runs of the GB1 Menu experiment. For Opus 4.7, the first \nPqNOpus{} campaigns of an earlier control run measure \nGbDistinctOpus{} and re-test \nGbShareOpus\%; Table~\ref{tab:landscapes} uses all \nGbNOpus{} campaigns of that run. Sol measures \nGbDistinctSol{} and Gemini
3.1 Pro \nGbDistinctPro. On GB1 Gemini 3.1 Pro loses mostly through exploration, \nGbExplPro, against \nGbSelPro{} at
selection.

\subsection{Sol on scrambled GB1}
\label{app:atwo}

Sol nominated the same candidate, true value \nFsrmTrue, in 12 of its 30 GB1 campaigns without the Menu. Its registered lead over the
guessing rule resamples campaigns by seed, so it counts those twelve campaigns as independent. Two checks count them as one: one resamples whole nominees, and the other counts the repeated nominee once. Both leave the estimate about the same and widen its interval to include zero (Table~\ref{tab:sollead}). Sol's lead on GB1 is therefore uncertain. The Menu's effect on its pick stays undetectable under both checks: \nSolGbMenuClu{} \nSolGbMenuCluci{} and \nSolGbMenuCol{}
\nSolGbMenuColci.

\begin{table}[tbp]
\centering
\small
\caption{Sol's pick minus the guessing rule's mean, with 95\% intervals that include the guessing rule's own
uncertainty: as registered, and under two checks that count its repeated nominee as one. The checks were not
registered.}
\label{tab:sollead}
\begin{tabular}{lcc}
\toprule
 & GB1 & PhoQ \\
\midrule
registered: campaigns resampled by seed & \estci{\nSolGbLeadReg}{\nSolGbLeadRegci} & \estci{\nPqSolLeadReg}{\nPqSolLeadRegci} \\
whole nominees resampled & \estci{\nSolGbLeadClu}{\nSolGbLeadCluci} & \estci{\nPqSolLeadNom}{\nPqSolLeadNomci} \\
repeated nominee counted once & \estci{\nSolGbLeadCol}{\nSolGbLeadColci} & \estci{\nPqSolLeadOnce}{\nPqSolLeadOnceci} \\
both checks together & -- & \estci{\nPqSolLeadOnceNom}{\nPqSolLeadOnceNomci} \\
\bottomrule
\end{tabular}
\end{table}

\subsection{How the Menu changed exploration and selection on the grid}
\label{app:split}

On the grid the Menu raised the best value Gemini 2.5 Flash and DeepSeek V3.1 measured, and it worsened the selection
of all three over-confirming models (Table~\ref{tab:split}). When a simple rule picks from the agents' own readings, the Menu improves the final pick of Opus 4.7 and Gemini 2.5 Flash. For DeepSeek V3.1 it makes no detectable difference.

\begin{table}[tbp]
\centering
\small
\caption{Change with the Menu on the grid, against each model's control on the same seeds, with 95\% intervals. For
Opus 4.7 only the selection loss was computed.}
\label{tab:split}
\begin{tabular}{lcc}
\toprule
 & best value measured & selection loss \\
\midrule
Gemini 2.5 Flash & \estci{\nSplitFlashExpl}{\nSplitFlashExplci} & \estci{\nSplitFlashSel}{\nSplitFlashSelci} \\
DeepSeek V3.1 & \estci{\nSplitDeepExpl}{\nSplitDeepExplci} & \estci{\nSplitDeepSel}{\nSplitDeepSelci} \\
Opus 4.7 & -- & \estci{\nSplitOpusSel}{\nSplitOpusSelci} \\
\bottomrule
\end{tabular}
\end{table}

\section{The five models on PhoQ}
\label{app:phoq}

The PhoQ runs repeat the GB1 control runs, those without the Menu, on scrambled PhoQ, for all five models. Each model ran at the settings of its GB1 runs, and only the landscape changed. The scramble is our own and differs from all of Wellocate's frozen
training, validation and test scrambles. Every model ran \nPqNOpus{} campaigns on the same seeds, so the models pair by
seed. The design, the outcomes and a named test were registered before the first call.

\paragraph{Re-testing on PhoQ.} On GB1 every model re-tests more than the best scripted rule. The registered test
extends that claim to PhoQ only if each model's 95\% interval lies above PhoQ's best scripted share. On PhoQ that
rule is the informed rule with no re-tests, so its share is zero. Every interval lies above it, and also above the
guessing rule's \nPqGuessShare\% (Table~\ref{tab:phoq}). Gemini 2.5 Flash lost \nPqFlashLostRounds{} slots a campaign
to rounds it could not complete, and a lost slot is never a re-test.

\paragraph{Picks against the guessing rule.} The registered comparison sets each pick against the guessing rule's mean
over its 1{,}500 reference seeds, \nPqGuessPick, and adds that rule's own uncertainty. A second, descriptive comparison
runs the guessing rule on each model's own seeds and pairs the two picks seed by seed. Paired this way, DeepSeek V3.1
and Gemini 3.1 Pro pick worse than the guessing rule on PhoQ, and the other three are about level with it. On GB1
every model except Sol picks worse. In the registered comparison DeepSeek V3.1, Gemini 3.1 Pro and Opus 4.7 pick worse, Gemini 2.5 Flash is level and Sol is better.

\begin{table}[tbp]
\centering
\small
\caption{The five models on scrambled PhoQ, with 95\% intervals. The last three columns are each model's pick minus the guessing rule's. The first sets it against that rule's mean over 1{,}500 reference seeds (the registered comparison). The other two pair it with the guessing rule on the model's own seeds, on PhoQ and on GB1.}
\label{tab:phoq}
\begin{tabular}{lccccc}
\toprule
 & & & \multicolumn{3}{c}{pick minus the guessing rule} \\
\cmidrule(lr){4-6}
 & share (\%) & final pick & registered & paired, PhoQ & paired, GB1 \\
\midrule
Opus 4.7 & \estci{\nPqShareOpus}{\nPqShareOpusci} & \estci{\nPqPickOpus}{\nPqPickOpusci} & \estci{\nPqLeadOpus}{\nPqLeadOpusci} & \estci{\nPairPqOpus}{\nPairPqOpusci} & \estci{\nPairGbOpus}{\nPairGbOpusci} \\
Gemini 2.5 Flash & \estci{\nPqShareFlash}{\nPqShareFlashci} & \estci{\nPqPickFlash}{\nPqPickFlashci} & \estci{\nPqLeadFlash}{\nPqLeadFlashci} & \estci{\nPairPqFlash}{\nPairPqFlashci} & \estci{\nPairGbFlash}{\nPairGbFlashci} \\
DeepSeek V3.1 & \estci{\nPqShareDeep}{\nPqShareDeepci} & \estci{\nPqPickDeep}{\nPqPickDeepci} & \estci{\nPqLeadDeep}{\nPqLeadDeepci} & \estci{\nPairPqDeep}{\nPairPqDeepci} & \estci{\nPairGbDeep}{\nPairGbDeepci} \\
GPT-6.1 Sol & \estci{\nPqShareSol}{\nPqShareSolci} & \estci{\nPqPickSol}{\nPqPickSolci} & \estci{\nPqLeadSol}{\nPqLeadSolci} & \estci{\nPairPqSol}{\nPairPqSolci} & \estci{\nPairGbSol}{\nPairGbSolci} \\
Gemini 3.1 Pro & \estci{\nPqSharePro}{\nPqShareProci} & \estci{\nPqPickPro}{\nPqPickProci} & \estci{\nPqLeadPro}{\nPqLeadProci} & \estci{\nPairPqPro}{\nPairPqProci} & \estci{\nPairGbPro}{\nPairGbProci} \\
\bottomrule
\end{tabular}
\end{table}

\paragraph{Sol's lead on PhoQ.} Sol nominated one candidate, IAQT, in \nPqTopSol{} of its campaigns, among
\nPqSolDistinctNominees{} different nominees. Its registered lead
therefore rests partly on that one candidate. Under the checks of Appendix~\ref{app:atwo} every interval includes zero
(Table~\ref{tab:sollead}), so Sol's lead on PhoQ is uncertain too.

\paragraph{A replication on other seeds.} An earlier run put the three earlier models on the same PhoQ landscape,
under the same Blind prompt, on other seeds. Its shares were
\nPqOamShareOpus\% \nPqOamShareOpusci{} for Opus 4.7, \nPqOamShareFlash\% \nPqOamShareFlashci{} for Gemini 2.5 Flash
and \nPqOamShareDeep\% \nPqOamShareDeepci{} for DeepSeek V3.1, close to the shares above. Its picks were
\nPqOamPickOpus, \nPqOamPickFlash{} and \nPqOamPickDeep, close to those above.

\section{The confirmation simulation}
\label{app:aone}

\paragraph{Rebuilding each campaign.} The nomination prompt holds a campaign's whole history, one line per order.
Replaying those orders through the simulated assay with the campaign's seed reproduces every reading. The noise and the failures are fixed by the seed, the candidate and its order count. A campaign was used only if every
logged outcome and reading was reproduced and the rebuilt registered pick and re-test count matched the original run's.
We rebuilt the grid campaigns of Gemini 3.1 Pro and Sol, and as a check those of Gemini 2.5 Flash and DeepSeek V3.1.
Every campaign passed.

\paragraph{Replacing the last round.} With the last round replaced, Gemini 3.1 Pro gains against its own pick at $m = 3$. It gains about as much at $m = 4$, where the interval just includes zero. At $m = 3$ its re-testing share rises to \nAoneProShareAfter\% (Table~\ref{tab:aone}). The registered rule applied
to its unchanged readings gains almost nothing, so the gain comes from the extra readings. Gemini 2.5 Flash and
DeepSeek V3.1 gain a little at $m = 1$. The simulation drops the agent's own final-round orders, so it gives up
whatever those would have found.

\begin{table}[!htbp]
\centering
\small
\caption{Change in final pick against each model's own pick in the confirmation simulation, with 95\% intervals.}
\label{tab:aone}
\begin{tabular}{llc}
\toprule
model & replay & change in final pick \\
\midrule
Gemini 3.1 Pro & last round, $m = 3$ & \estci{\nAoneProKthree}{\nAoneProKthreeci} \\
Gemini 3.1 Pro & last round, $m = 4$ & \estci{\nAoneProKfour}{\nAoneProKfourci} \\
Gemini 3.1 Pro & registered rule on unchanged readings & \estci{\nAoneProRepick}{\nAoneProRepickci} \\
Gemini 2.5 Flash & last round, $m = 1$ & \estci{\nAoneFlashKone}{\nAoneFlashKoneci} \\
DeepSeek V3.1 & last round, $m = 1$ & \estci{\nAoneDeepKone}{\nAoneDeepKoneci} \\
Gemini 3.1 Pro & last two rounds, $m = 3$ & \estci{\nAtwoProKthree}{\nAtwoProKthreeci} \\
GPT-6.1 Sol & last two rounds, $m = 1$ & \estci{\nAtwoSolKone}{\nAtwoSolKoneci} \\
\bottomrule
\end{tabular}
\end{table}

\paragraph{The last two rounds.} Replacing the last two rounds, with the top $m$ chosen after the round before them,
recovers about \nAtwoProRecoveredOwn\% of Gemini 3.1 Pro's selection loss at $m = 3$. One round recovers \nAoneProRecoveredKthree\% at the same $m$. Sol's pick falls slightly at $m = 1$, with an interval that includes zero. This run was not planned in advance.

\section{The pre-registered prompt study}
\label{app:grid}

This appendix reports the pre-registered prompt study: three models under five prompt conditions on the
grid, with its prompt checks, its zero-noise run and its independent check. It is the evidence behind
the over-confirmation in Section~\ref{sec:both}. Table~\ref{tab:cells} compares each cell with the guessing rule on
how good the best candidate it tested was and on how well it picked among what it tested.

\subsection{Setup}
\label{sec:envgrid}
\label{sec:design}

The grid uses the generated landscape, campaign, noise and failures of Section~\ref{sec:env} and
Appendix~\ref{app:env}. The agent is told only that it is running a numerical optimisation campaign, and never that
it is working on proteins.

\paragraph{Why the landscape is generated.} A generated landscape makes the true value of every candidate
known, and no model can have seen its best candidate. A published landscape cannot guarantee either. Given the GB1
protein landscape of \citet{wu2016adaptation} with its real amino-acid letters, Opus 4.7 named the best
measured variant in round one in 11 of 20 campaigns. Four random viable picks would do so 0.035\% of the
time. The prompt study uses the grid only, and the five-model study adds GB1 and PhoQ.

\paragraph{Five prompt conditions.} Five conditions vary what the agent is told, each adding to the one
before. \textbf{Blind} says nothing about measurement error. \textbf{Quantitative} adds that readings carry
error and gives $\sigma(f)$. \textbf{Mechanism} adds that averaging $n$ readings reduces the standard error
by $\sqrt{n}$, and permits re-ordering, without recommending it. \textbf{Aggregated} keeps Mechanism's text
and changes the environment: the history is shown as a count, mean and standard error per
candidate, in place of one line per reading. \textbf{Decision-rule} adds the selection rule outright. No
condition ever instructs the agent to re-test, because that would measure instruction-following.

\begin{sloppypar}
\paragraph{Models and scale.} Three models, called Opus 4.7, Gemini 2.5 Flash and DeepSeek V3.1
throughout this paper and requested from OpenRouter as \texttt{anthropic/claude-opus-4.7},
\texttt{google/gemini-2.5-flash} and \texttt{deepseek/deepseek-chat-v3.1}, at temperature 0. We call one
model under one prompt condition a \emph{cell}, so the prompt study has 15 cells. It holds
5{,}075 campaigns and 162{,}986 model calls. Cells are sized per model: 695, 139 and 181 campaigns per cell.
\end{sloppypar}

\paragraph{Pre-registration.} The design, the outcome, the six confirmatory contrasts and the analysis were
fixed before the prompt study ran (Appendix~\ref{app:amendments}). The six contrasts are Aggregated against Mechanism and
Decision-rule against Mechanism, for each model, with Holm correction for multiple comparisons across the six.
Contrasts are paired by seed. The seed fixes measurement noise and failures, so at a given seed the $k$-th reading
of a candidate is the same in every condition.

\paragraph{The scripted rules.} The guessing and informed rules of Appendix~\ref{app:refs} are the grid's
references. Both do best at $k^\ast = 1$, one slot in four.

\subsection{Re-testing share}
\label{sec:results}

\begin{figure}[tbp]
\centering
\includegraphics[width=\textwidth]{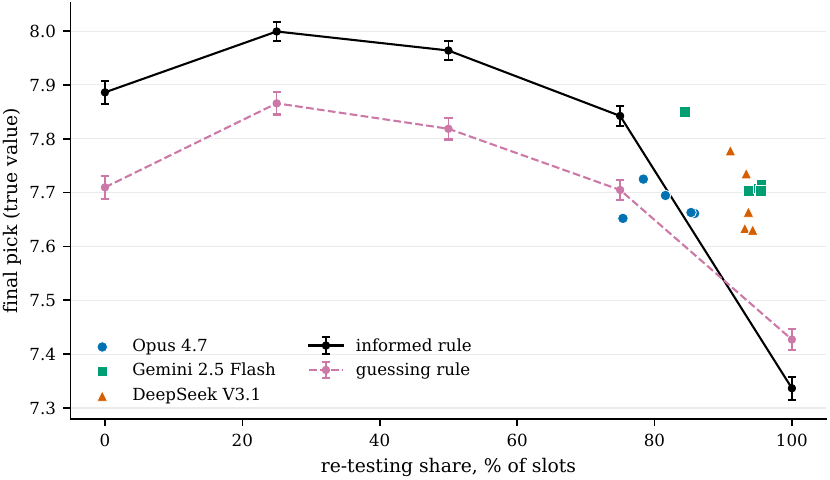}
\caption{True value of the final pick against the re-testing share, one point per cell. The two lines are the
informed rule (upper, told which candidates are viable) and the guessing rule (lower), spending $k=0$ to 4 of the
four slots each round re-testing their current best, with 95\% intervals.}
\label{fig:gridheadline}
\end{figure}

Agents spend most of their measurements re-testing candidates they have already measured. Across the 15
cells the re-testing share runs from 75\% to 96\% of used slots; both scripted rules spend a
quarter (Table~\ref{tab:cells}, Figure~\ref{fig:gridheadline}). The count includes re-orders of the reference
candidate, whose score the agent is given at the start. On this count every cell re-tests more than a
scripted protocol that measures each candidate three times. Without those re-orders the share is 57\% to
75\%, and six cells fall below that protocol. Counting only slots spent on a candidate that already has a successful reading leaves out re-orders after a failed order. That count gives 74.7\% to 95.1\%, within 1.4 points of the headline
in every cell.

\paragraph{The scripted lab protocols.} Three scripted protocols spend the same 60 slots the way a wet lab
is often run, searching around their current best as the scripted rules do. Triplicate-everything measures 20 candidates three times,
66.7\% of slots on repeats. Duplicate-everything measures 30 twice, 50.0\%. Screen-then-confirm takes 48
single readings and then confirms the leaders, 20.0\%. These protocols are simulated, and a study with human
experimenters is future work.

\begin{table}[tbp]
\centering
\small
\caption{Results for each cell, one model under one prompt condition. \emph{Share} is the re-testing share, the fraction of used slots spent re-testing a candidate already ordered. \emph{Excl.\ start} leaves out re-orders of the start, whose reading the agent is given in advance. \emph{Distinct} is the mean number of different candidates measured per campaign, and \emph{pick} the mean true value of the final pick. The last three columns compare each cell with the guessing rule at its best setting, which measures 28.94 distinct candidates and reaches 7.87. \emph{Tested} is how much further the best candidate the agents measured falls short of the global best. \emph{Picked} is the change in how far their pick falls short of the best they measured, negative where they picked better. \emph{Net} is the sum, the agents' shortfall in the true value of the final pick; rounded parts may not add to it exactly.}
\label{tab:cells}
\begin{tabular}{llrrrrrrr}
\toprule
 & & \multicolumn{2}{c}{share (\%)} & & & \multicolumn{3}{c}{against the guessing rule} \\
\cmidrule(lr){3-4}\cmidrule(lr){7-9}
model & condition & all & excl.\ start & distinct & pick & tested & picked & net \\
\midrule
Opus 4.7 & Blind & 75.4 & 62.8 & 11.3 & 7.65 & $+0.33$ & $-0.12$ & $+0.21$ \\
 & Quantitative & 78.4 & 65.5 & 10.4 & 7.73 & $+0.25$ & $-0.11$ & $+0.14$ \\
 & Mechanism & 81.6 & 68.1 & 9.0 & 7.69 & $+0.31$ & $-0.14$ & $+0.17$ \\
 & Aggregated & 85.8 & 70.4 & 7.0 & 7.66 & $+0.37$ & $-0.17$ & $+0.20$ \\
 & Decision-rule & 85.3 & 70.8 & 7.3 & 7.66 & $+0.36$ & $-0.16$ & $+0.20$ \\
\addlinespace
Gemini 2.5 Flash & Blind & 93.8 & 69.5 & 3.3 & 7.70 & $+0.37$ & $-0.21$ & $+0.16$ \\
 & Quantitative & 84.4 & 67.3 & 7.7 & 7.85 & $+0.18$ & $-0.16$ & $+0.02$ \\
 & Mechanism & 95.2 & 73.3 & 2.7 & 7.71 & $+0.38$ & $-0.22$ & $+0.16$ \\
 & Aggregated & 95.6 & 74.6 & 2.6 & 7.71 & $+0.37$ & $-0.22$ & $+0.15$ \\
 & Decision-rule & 95.5 & 73.8 & 2.6 & 7.70 & $+0.38$ & $-0.21$ & $+0.16$ \\
\addlinespace
DeepSeek V3.1 & Blind & 91.0 & 68.6 & 4.9 & 7.78 & $+0.26$ & $-0.17$ & $+0.09$ \\
 & Quantitative & 93.1 & 63.1 & 3.9 & 7.63 & $+0.41$ & $-0.18$ & $+0.23$ \\
 & Mechanism & 93.7 & 56.9 & 3.7 & 7.66 & $+0.38$ & $-0.18$ & $+0.20$ \\
 & Aggregated & 94.3 & 57.0 & 3.4 & 7.63 & $+0.43$ & $-0.19$ & $+0.24$ \\
 & Decision-rule & 93.4 & 58.8 & 3.8 & 7.74 & $+0.31$ & $-0.18$ & $+0.13$ \\
\bottomrule
\end{tabular}
\end{table}

\begin{figure}[tbp]
\centering
\includegraphics[width=\textwidth]{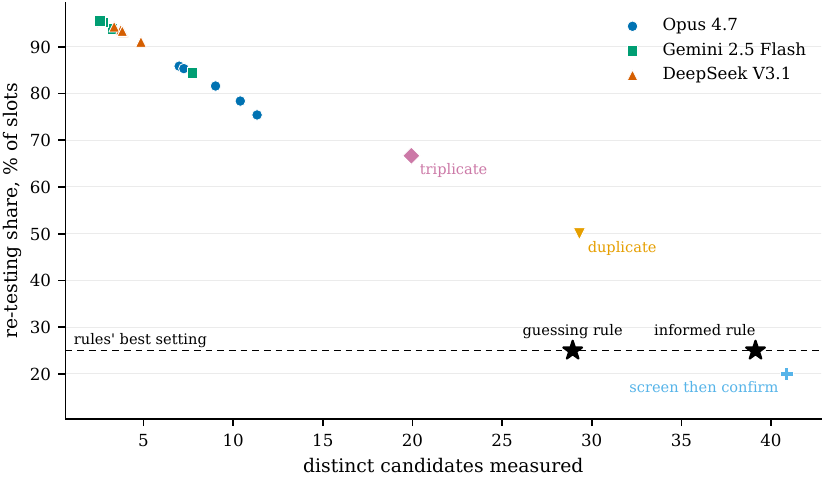}
\caption{Re-testing share against distinct candidates measured, one point per cell, with the informed and
guessing rules at their best setting and the three scripted lab protocols.}
\label{fig:gridsplit}
\end{figure}

The re-tests pay off in selection. Agents picked well among the candidates they had measured, closer to the best of
what they tested than the guessing rule. The best candidate they tested was much worse than that rule's best, because they tested so few (Figure~\ref{fig:gridsplit}). In every cell the shortfall from testing too few outweighs
the gain from picking well, so the agents end up behind (Table~\ref{tab:cells}, Figure~\ref{fig:shortfall_global}; Appendix~\ref{app:further}).
Their stated reasons show that they climb from their current best one letter at a time, as the scripted rules do,
in lines such as \emph{``vary pos4, C works
so try neighbor''}. They climb a few steps and then spend most of the budget confirming the same few candidates, so
the climb stalls.

\subsection{Exact readings}

\begin{figure}[tbp]
\centering
\includegraphics[width=\textwidth]{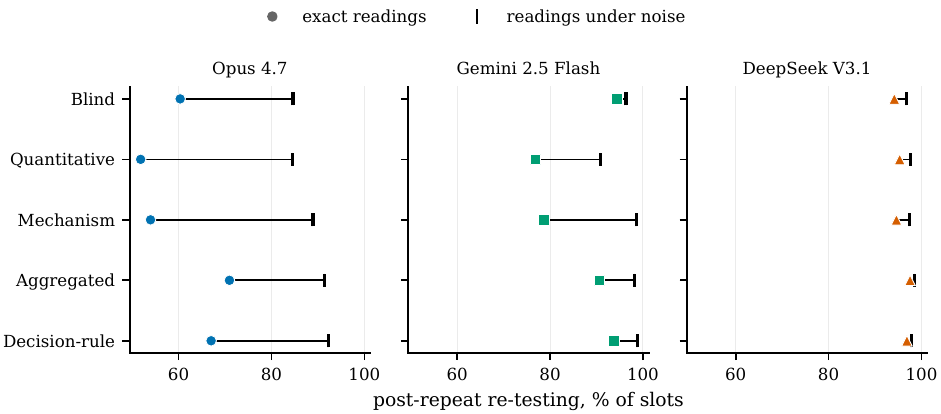}
\caption{Re-testing after the first repeat when every reading is exact (the marker) and the level implied under
noise (the bar), one row per cell. A repeat of an exactly known value returns no new information.}
\label{fig:zero}
\end{figure}

Removing the noise lowers re-testing in every cell, significantly in 12 of 15. Re-testing still takes roughly half
to nearly all of the slots, and with exact readings no re-test is needed. The zero-noise run sets $\sigma$ to zero and changes nothing
else, so the agent is not told its readings are exact; Appendix~\ref{sec:indep} tests the case where it is
told. Here re-testing is counted after the first repeat an agent sees, as the share of slots spent on a candidate
that already has a successful reading. This is a stricter count than the headline share, and it is compared with the
same count under noise at the same seeds (Figure~\ref{fig:zero}).

At $\sigma=0$, re-testing after the first repeat runs from 51.8\% to 97.5\% across
the 15 cells, with a median of 90.6\%. The lowest cell is Quantitative with Opus 4.7, at 51.8\%, 95\%
interval [48.2\%, 55.4\%]. The highest is Aggregated with DeepSeek V3.1, at 97.5\%, [96.7\%, 98.3\%]. By
model, Opus 4.7 runs 51.8\% to 71.0\%, Gemini 2.5 Flash 76.8\% to 94.4\% and DeepSeek V3.1 94.1\% to
97.5\%. All 15 paired differences from the same cells with noise are negative, from $-35.0$ to $-0.9$ percentage points. Of these, 12 are significant at 0.05 after Holm correction within that run. Every cell pairs all 40 seeds. The size of the drop tracks the level a model starts from. Opus 4.7 falls furthest. DeepSeek V3.1 falls least, from levels of 96.7\% to 98.5\% under noise. Elsewhere in these appendices, zero-noise numbers count all used slots.

\subsection{Independent check}
\label{sec:indep}
\label{app:indep}

In a separate one-round test with exact readings, the models still spent from about a quarter to over half of the round re-testing. Adding noise raised re-testing sharply, so the models do respond to noise. The test was
written separately from Wellocate, with its own world generator, prompt and scorer, and shares no code with it.

Each call shows one fixed round of an optimisation task. Every candidate is valid, so nothing is ever rejected. A
seven-round history is already measured, and the best candidate has already been measured twice. Three versions vary only what is true and what is said about measurement error. They are exact readings with exactness stated, exact
readings with nothing said, and noisy readings with the noise stated. Thirty paired histories per version, three
models, temperature 0, 270 calls. Each reply is scored over four slots. A missing line counts as a new candidate, and so does a name with letters outside the alphabet.

\begin{table}[h]
\centering
\small
\caption{Share of the four chosen measurements that repeat an already-measured candidate.}
\begin{tabular}{lrrrr}
\toprule
model & exact, stated & exact, unstated & noisy, stated & noise effect (paired) \\
\midrule
Opus 4.7         & 24.2\% & 24.2\% & 61.7\% & $+37.5$ pp (SE 5.3) \\
Gemini 2.5 Flash & 45.0\% & 50.8\% & 95.0\% & $+50.0$ pp (SE 7.0) \\
DeepSeek V3.1    & 55.8\% & 49.2\% & 94.2\% & $+38.3$ pp (SE 8.3) \\
\bottomrule
\end{tabular}
\end{table}

A reading rule was fixed before any call. A repeat share of 25\% or more in the exact-stated version confirms
the behaviour; 5\% or less attributes it to our environment; anything between is partial.
Gemini 2.5 Flash and DeepSeek V3.1 clear the 25\% line. Opus 4.7 at 24.2\% falls just
below it, so two models give a clear reproduction and one is partial.

Stating that readings are exact made no detectable difference. Paired by history, nothing said minus exactness stated is $+0.0$ pp (SE 2.4) for Opus 4.7, $+5.8$ (SE 7.1) for Gemini 2.5 Flash and $-6.7$
(SE 6.9) for DeepSeek V3.1. In the zero-noise run the noise effect looks small for Gemini 2.5 Flash and DeepSeek V3.1, because their re-testing is already high with exact readings. Averaged over the five conditions, they re-test 81.6\% and 89.6\% of all used slots there, against 51.4\% for Opus 4.7.

\subsection{Pre-registered contrasts}

If the behaviour came from a history that is hard to read, or from a missing rule, fixing those should reduce
it. Six contrasts were fixed before the prompt study ran, comparing Aggregated and Decision-rule with Mechanism for
each model, with Holm correction across the six (Table~\ref{tab:confirm}). One survives the correction: giving the selection rule outright improved the final pick for DeepSeek V3.1,
by $+0.0699$. It did not hold up: a pre-registered replication on 120 new seeds gave $+0.008$
$[-0.036, +0.052]$ (Appendix~\ref{app:replication}). In the original run the gain came through what DeepSeek V3.1 measured. Against the guessing rule, its
shortfall in the best candidate tested fell from 0.38 to 0.31, and its selection gain stayed at 0.18. It picked among what it tested as well as before (Table~\ref{tab:cells}). It
picked the highest-average candidate about as often with the rule as without it (86\% against 90\%; an exact
McNemar test for paired choices gives $p = 0.36$). Its re-testing share moved by less than half a point (93.7\% against 93.4\%), and it tried
about as many candidates (3.7 against 3.8). Opus 4.7 responded differently. The rule raised its
highest-average picks from 74\% to 86\% ($p < 0.001$), and it also re-tested more and tried about two fewer
candidates. Opus 4.7 was nominally worse under both interventions; its unadjusted intervals exclude zero, and both differences lose significance after correction for the six contrasts. Gemini 2.5 Flash showed no detectable change.
Arithmetic alone does not explain the habit either. The Aggregated condition shows each candidate's count, mean and
standard error already computed, and re-testing was slightly higher under it than under Mechanism for all three
models. We do not read any of this as evidence that stating a decision rule helps agents.

\begin{table}[tbp]
\centering
\small
\caption{The six confirmatory contrasts on the final pick's true value, paired by seed, with Holm correction
across the family. The primary $p$ comes from a bootstrap that resamples seeds (studentised wild cluster bootstrap, 20{,}000 draws). The intervals are unadjusted and the Holm $p$ adjusts for the family of six. Seed
counts differ because each model's cells were sized separately (Section~\ref{sec:design}). $^\dagger$Did not hold up on 120 new seeds: $+0.008$, 95\% bootstrap interval $[-0.036, +0.052]$
(Appendix~\ref{app:replication}).}
\label{tab:confirm}
\begin{tabular}{llrrrr}
\toprule
model & contrast & seeds & diff & 95\% CI & Holm $p$ \\
\midrule
Opus 4.7     & Aggregated    & 695 & $-0.0334$ & $[-0.0606, -0.0061]$ & 0.0873 \\
Opus 4.7     & Decision-rule & 695 & $-0.0316$ & $[-0.0583, -0.0049]$ & 0.0873 \\
DeepSeek V3.1& Aggregated    & 181 & $-0.0332$ & $[-0.0757, +0.0092]$ & 0.3721 \\
DeepSeek V3.1& Decision-rule & 180 & $+0.0699$ & $[+0.0271, +0.1126]$ & \textbf{0.0075}$^\dagger$ \\
Gemini 2.5 Flash & Aggregated    & 139 & $+0.0067$ & $[-0.0130, +0.0265]$ & 1.0000 \\
Gemini 2.5 Flash & Decision-rule & 139 & $-0.0044$ & $[-0.0278, +0.0189]$ & 1.0000 \\
\bottomrule
\end{tabular}
\end{table}

\subsection{Prompt checks}
\label{sec:prompt}
\label{app:prompt}

A review of the prompts found two flaws. The round and replacement prompts did not say that the final pick is
scored on its true value. The replacement prompt also stated one order per round, where a round has four. A
wrong guess about viability could cost a slot, so rejection risk might also push agents toward re-testing
candidates known to work.

Three prompt changes were tested, each absent from the original prompts. The \emph{goal} option adds one sentence to the goal in the round and replacement prompts, the same in all five conditions. It reads ``At the
end you will nominate one candidate, and your score is that candidate's true value.'' It says nothing about
readings, because the Blind condition withholds that readings carry error. The nomination prompt was left as it
was, since it already asks for the highest true score. The \emph{budget} option makes the replacement prompt's budget line
read ``15 rounds, 4 orders per round, 60 measurements in total.'' It had stated one order per round.

The \emph{free-rejection} option makes a rejected order free. In every prompt, the sentence on replacements and lost slots becomes
``An order for a non-viable candidate is rejected at no cost and you choose again. No slot is ever lost to a
non-viable order.'' The agent may then be asked again up to 20 times for one slot. A slot that reaches that cap
is lost and counted, and no slot in check B reached it. The guessing rule reached it in 353 of
its 1{,}500 campaigns at one re-test per round, and still lost 0.86 slots per campaign.

Check A used the goal and budget options and check B all three, each on Blind and Mechanism for all three models. Both ran on 60 seeds, each paired with the original campaign on the same seed (Table~\ref{tab:prompt}). A rerun of the original prompts ran on the same day as check A. The registered statement for check B reads ``rejection risk does not produce the habit''. Intervals are 95\% percentile bootstraps over the paired
campaigns.

Check A fixes both flaws. With it, the re-testing share moved by less than 1.1 points in five of six cells. For Gemini 2.5 Flash under Blind it fell by 4.6 points [$-6.6$, $-2.6$],
and under Mechanism Gemini's final pick fell by 0.160 [$-0.256$, $-0.070$]. The registered margin for final picks,
$\pm 0.06$, could not be met with 60 pairs in four cells even with no true change. The re-testing share is
therefore the informative outcome. The same-day rerun of the original prompts moved the re-testing share by at most 1.1 points from the original campaigns in every cell. The two Gemini shifts remain against this
control, so they come from the wording.

\begin{table}[tbp]
\centering
\small
\caption{Re-testing share in the prompt checks, as a percentage of used slots, on the same 60 seeds in every column. \emph{Grid} is the original campaigns on those seeds. Check~A states that the final pick is scored on its true value and corrects the replacement prompt's order count. Check~B also makes a rejected order free. \emph{Same-day rerun minus grid} is the change, in points, for the original prompts run again on the same day as check~A.}
\label{tab:prompt}
\begin{tabular}{llrrrr}
\toprule
model & condition & grid & check A & check B & same-day rerun minus grid \\
\midrule
Opus 4.7 & Blind & 76.0 & 76.6 & 74.8 & $-1.1$ \\
 & Mechanism & 82.1 & 81.1 & 80.9 & $+0.2$ \\
\addlinespace
Gemini 2.5 Flash & Blind & 94.4 & 89.8 & 83.3 & $-0.1$ \\
 & Mechanism & 95.2 & 95.1 & 89.0 & $-0.9$ \\
\addlinespace
DeepSeek V3.1 & Blind & 90.8 & 90.4 & 90.0 & $-0.9$ \\
 & Mechanism & 93.6 & 93.6 & 93.5 & $+0.1$ \\
\bottomrule
\end{tabular}
\end{table}

In check B every cell still spent 74.8\% or more of its slots on re-tests, against 23.96\% for
the guessing rule at its best setting under the same rule. Gemini 2.5 Flash re-tested 6 to 11 points less and picked better, and it stayed far above the
guessing rule. The final pick of Opus 4.7 under Mechanism fell by 0.1340 [$-0.2324$, $-0.0347$].

\subsection{Exploration, final picks and rejections}
\label{app:further}

\begin{figure}[tbp]
\centering
\includegraphics[width=\textwidth]{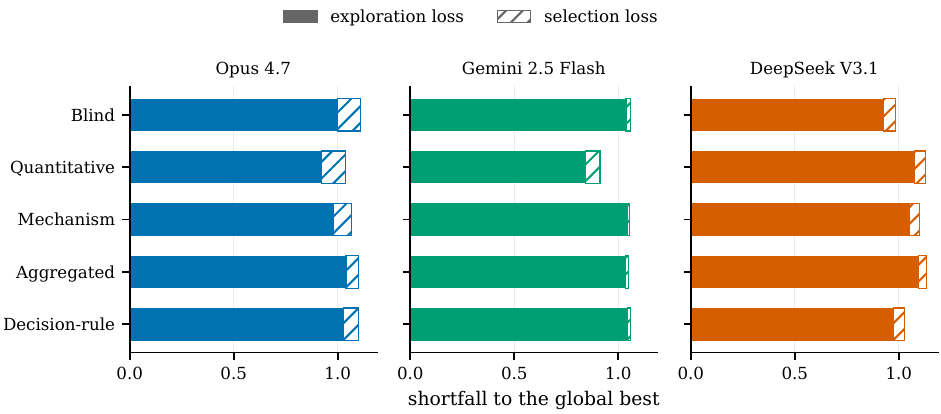}
\caption{The shortfall to the global best in each cell, split into exploration loss and selection loss, which
sum to the whole. The guessing rule's own split, 0.67 to exploration and 0.23 to selection, is the reference point of
Figure~\ref{fig:loss}.}
\label{fig:shortfall_global}
\end{figure}

\paragraph{Prompt conditions and exploration.} The explanatory conditions raise re-testing and cut
exploration. For Opus 4.7, distinct candidates fall from 11.34 under Blind to 9.02 under Mechanism and 6.99
under Aggregated, and its re-testing share rises from 75.4\% to 81.6\% and then 85.8\%. This is
descriptive across conditions and is not one of the pre-registered contrasts.

\paragraph{Final picks against the scripted rules.} Cell means of the final pick's true value run from
7.6308 to 7.8501. At $k^\ast$ the guessing rule reaches 7.8659 and the informed rule 7.9993. Against the scripted rules with no re-tests, the answer depends on the rule. Five of the 15 cells sit above the guessing
rule's $k{=}0$ mean of 7.7098, two with intervals entirely above, and every cell sits below the informed
rule's 7.8862. The most exploratory cell reaches 39.2\% of the guessing rule's count of distinct candidates. Split
against the global best, exploration accounts for 0.8428 to 1.0928 of the shortfall and selection for 0.0087 to 0.1148. In every cell exploration is thus 88.9\% to 99.2\% of the total and selection 0.8\% to 11.1\% (Figure~\ref{fig:shortfall_global}).

\paragraph{Rejections.} Cell means run 2.57 to 11.34 distinct candidates, against 28.94 for the guessing rule and
39.14 for the informed rule. Over all 5{,}075 prompt-study campaigns the agents were rejected 16.75 times per campaign and
lost 0.71 slots, against 83.02 and 11.70 for the guessing rule under the same rules. The agents were rejected less
often per campaign because they ordered far fewer new names: most of their slots went on re-tests. Per new name they
were rejected 72.2\% of the time, against 74.4\% for the guessing rule. New names are rejected orders plus
candidates measured, leaving out the starting candidate both are given. So the agents pick viable names about as
well as the guessing rule's neighbour sampling. Across the 15 cells the rate runs from 68.5\% to 89.0\%. Eight
of the 15 cells are rejected more often than the guessing rule and seven less often. The eight are Gemini 2.5
Flash in every condition except Quantitative, and DeepSeek V3.1 in every condition except Decision-rule.

\section{Temperature, reasoning and a biology framing}
\label{app:panels}

\paragraph{Temperature.} Three models under Mechanism at temperature 1.0, on the grid's own seeds. Re-testing moves by $-0.3$
percentage points (SE 0.8) for Opus 4.7 and $-1.0$ (SE 0.3) for DeepSeek V3.1. It falls 5.5 (SE 1.0) for Gemini 2.5 Flash, which leaves it at 89.9\% of slots. The final pick
moves by at most 0.07 in either direction, with standard errors of 0.04 to 0.06.

\paragraph{Reasoning.}
\label{app:reasoning}
This experiment runs two of the five conditions on two of the three models, at reasoning effort medium. It is paired by seed against the prompt study, with 35 to 40 seeds per cell after exclusions. Holm correction runs across
its four cells only.

Reasoning shifted agents from re-testing toward exploration and their picks worsened, mainly because they
found worse candidates. The re-testing share falls by 7.5 to 12.4 percentage points and distinct candidates
rise by 3.7 to 7.1 per campaign in all four cells. The final pick's true value falls by 0.099 to 0.263,
and all four cells reject at 0.05 after Holm. The true value of the best candidate the agent measured drops
from 7.6854--7.7387 in the prompt study to 7.5302--7.6253 under reasoning. Exploration loss rises by $+0.06$ to $+0.19$, and selection loss by only $+0.007$ to $+0.085$. In 9 of 80 Gemini 2.5 Flash reasoning campaigns the model returned no final answer, eight of them at the 6{,}000-token output limit, and these are excluded. Keeping the nine excluded campaigns
leaves the comparison unchanged: with all 40 paired seeds the differences are $-0.0987$, $-0.2528$,
$-0.2016$ and $-0.2338$, and all four still reject.

\paragraph{Biology framing.}
\label{app:bio}
Each of the three models ran three cells of 40 campaigns on the Blind prompt, each with one added sentence that gives the agent a wet-lab role. The cells hold the role alone, the role with exact readings, and the role with one more sentence saying readings are noisy. The result covers only what one sentence of lab context does to the behaviour. The candidates are four letters from A to T. Seventeen of the twenty letters are amino-acid codes. The landscape is generated, so a biochemical intuition the framing evokes has nothing to attach to.

The role lowers re-testing in all three models against the same prompt without it. The changes are $-6.5$ percentage points
for Opus 4.7 $[-8.5, -4.5]$, $-9.8$ for Gemini 2.5 Flash $[-12.7, -7.0]$ and $-2.9$ for
DeepSeek V3.1 $[-4.7, -1.2]$. Re-testing persists under the role with exact readings, at 49.1\% to
86.4\% of slots. The same persistence appears with no role at all. The zero-noise run's own Blind cells, which carry the
prompt these role cells are built on, sit at 52.6\%, 92.9\% and 88.0\% of all used slots. Within this experiment, exact readings lower re-testing against the role cell by 19.8, 8.3 and 1.8 percentage points, paired by
seed, with DeepSeek V3.1's interval including zero. Saying that readings are noisy raised re-testing a little against the role cell, in all three models.

\section{A pre-registered replication of the Decision-rule result}
\label{app:replication}

One confirmatory contrast survived Holm correction: DeepSeek V3.1's final picks were better under
Decision-rule than under Mechanism. We ran that comparison again on 120 seeds that no earlier run had used.
The design, the outcome and the reading rule were fixed before any call. The landscape, the prompts and
the settings were those of the prompt study.

The outcome is the paired difference in the final pick's true value, Decision-rule minus Mechanism. One
Decision-rule campaign whose nomination did not parse is excluded, as in the prompt study's analysis, which leaves 119
pairs. The rule, fixed in advance, has three readings. A positive mean whose 95\% bootstrap interval excludes
zero replicates. A positive mean whose interval includes zero reads as ``direction holds, not significant''.
A mean at or below zero does not replicate.

The mean difference is $+0.008$, with a 95\% bootstrap interval of $[-0.036, +0.052]$. By the rule, the
direction holds and the difference is not significant. Of the 119 pairs, 37 improved, 33 got worse and 49
were unchanged. The interval's upper end lies below the original estimate of $+0.0699$ and below 0.06, the smallest
effect the study was designed to detect.

Not pre-registered: because the landscape and settings are the same, the replication's conditions can be set against the original run's. The Decision-rule means match: 7.736 originally and 7.736 in the replication. The
Mechanism mean rose from 7.664 to 7.728, a difference of $+0.064$ (SE 0.023). This is consistent with the
original gap coming mostly from a Mechanism sample that happened to score low.

\section{The learnability simulation}
\label{app:sim}

This simulation asks whether the best number of re-tests can be learned from the environment's reward alone. It makes no model calls. It runs the informed and the guessing rule at $K \in \{1,2,3\}$ and at
noise of 0.5, 1 and 2 times ours, 18 settings in all. Each uses 200 fresh landscapes, one per seed, and training and
held-out landscapes come from separate seed ranges.

Trained on the environment's reward alone, a two-parameter learner that chooses the number of re-tests and the
selection rule settles on $k=1$ with the registered selection rule of Section~\ref{sec:under}. That is $k^\ast$ or tied with $k^\ast$ in all 18
simulated settings. Sweeping $k$ gives $k^\ast = 1$ in 13 settings, $k^\ast = 2$ in 3 and $k^\ast = 0$ in 2. In 16 of the 18 the
gap between $k^\ast$ and its runner-up sits inside two standard errors, so most settings are ties. Raw paired
differences between the policy at $k^\ast$ with the registered selection rule and the rule that never re-tests run $-0.0077$ to $+0.2231$
on the held-out landscapes. Both negative values are at half noise.

\begin{figure}[tbp]
\centering
\includegraphics[width=\textwidth]{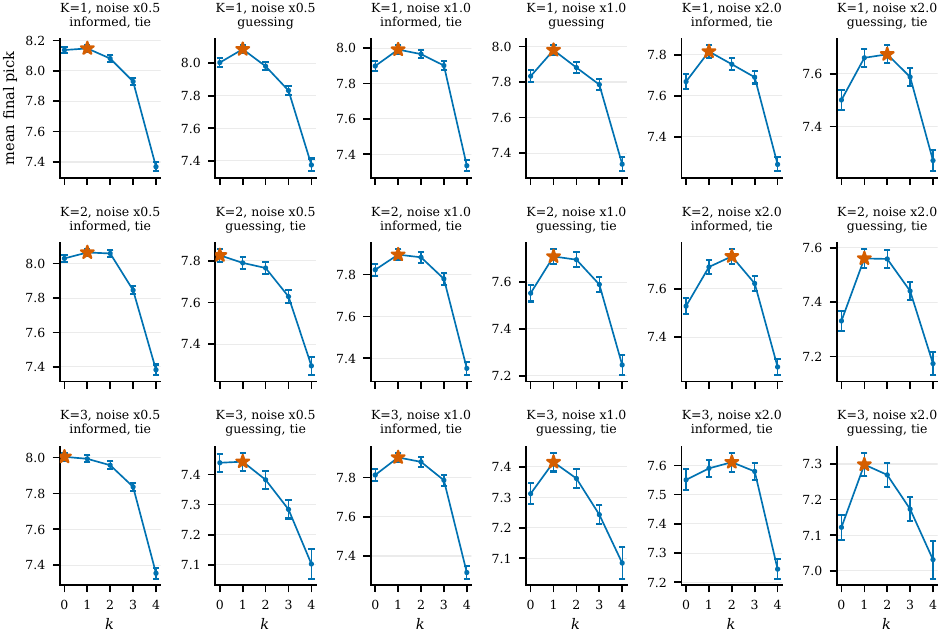}
\caption{The learnability simulation: mean final pick against $k$, the slots per round spent re-testing the current best, with standard errors. Each panel is one setting of ruggedness $K$, noise (as a multiple of assay noise) and rule, over 200 generated landscapes. The star marks the best $k$. A panel marked \emph{tie} is one where the paired gap
to the runner-up is inside two of its own standard errors.}
\label{fig:sim}
\end{figure}

The right rate depends on the noise. It is zero when readings are exact. In the simulation one repeat per round is best
or tied in all 18 settings. At half noise, never re-testing ties with one repeat in two settings at 200
landscapes, and at 1{,}500 landscapes the best is never zero. 

\paragraph{The registered score and collapsed campaigns.} The pre-registered score rescales each result between those of
two scripted endpoints. Across landscapes those endpoints move by more than the gap between them. A held-out score rescaled with training endpoints is therefore unstable: 0.122 to 7.765 across the 16 cells where it is defined.
Rescored with endpoints from the same held-out landscapes it runs 0.784 to 1.000. It is exactly 1.000 in the 11 cells
where $k^\ast = 1$ and the score is defined, because the learner selects that policy. It is undefined in four cells at
half noise, where the held-out span is zero or negative. No claim in this paper depends on it. The guessing rule's
means at high re-testing are over surviving campaigns, since a campaign collapses when fewer than two candidates are measured. At $K=3$, 138 of 200 campaigns collapse at $k=4$ and 12 of 200 at $k=3$.

\paragraph{The larger run.} At 1{,}500 landscapes per setting the picture holds: one
re-test a round is best or tied in all 18 settings. The three settings where two came out ahead
are all at double noise and all within two standard errors. Rescored with held-out endpoints, the learner
selects one re-test a round with the registered selection rule in every setting. It scores exactly 1.000 in the 15 settings where
$k^\ast = 1$ and between 0.819 and 0.925 in the other three. Figure~\ref{fig:sim} is drawn at 200 landscapes,
the pre-registered primary, since only that run recorded a standard error for each $k$.

\section{Training a policy in Wellocate}
\label{app:b11}

Trained on Wellocate's reward alone, a two-parameter rule already settles on a setting that is best or tied for
best on generated landscapes (Appendix~\ref{app:sim}). This appendix trains a full policy.

\paragraph{Design.} We trained a small policy with proximal policy optimisation (PPO) through Wellocate's released
interface. It sees what a language model agent sees: the readings, the outcomes and the remaining budget, and never a
true value or the set of viable candidates. At each step it scores a set of candidates. The set holds re-tests of
what it has ordered, every single-letter change to the start and to its eight best candidates, and one random untried
candidate. A second network picks the nominee and a third estimates value. Its inputs are scaled to the campaign's own
readings, so the same networks run on any landscape. Run A trains on generated grid landscapes only. Run B also
trains on scrambled GB1 and is tested on PhoQ. Each run has three seeds and keeps the checkpoint with the best score
on its validation split. The design, the test and the success criteria were registered before training. The policy had to beat both its own untrained
initialisation and random ordering, each with a 95\% interval above zero and every seed positive. It had to do so on
the grid test split to show the task is learnable, and on the scrambled GB1 and PhoQ test splits to show transfer.

\paragraph{The registered result.} Run A meets both criteria (Table~\ref{tab:b11reg}). Run B beats its initialisation,
and its interval against random ordering includes zero, so it fails the transfer criterion. A rerun decided after the
test, with an exploration bonus five times the registered one, meets it on PhoQ. It beats random ordering by
\nPolEntRandPq{} \nPolEntRandPqci{} with every seed positive (Table~\ref{tab:b11phase2}). The registered intervals
resample episodes only. Table~\ref{tab:b11reg} gives every interval from a two-level bootstrap that resamples
landscapes and then episodes; this check is post-hoc. Run A's criteria all hold under it, and also under an interval
across the three seeds. Under the two-level bootstrap, two comparisons with the guessing rule include zero, on the grid and on PhoQ, so we make no
claim that the policy beats the guessing rule. Five more seeds of Run A, trained and tested the same way, replicate
this: every seed beats its initialisation and random ordering on all three test splits (Table~\ref{tab:b11phase2}).

\paragraph{Robustness.} Everything from here on was decided after the registered test. Run A plays every condition without retraining, with
the networks of the registered test. Its lead over random ordering holds when the noise is halved or doubled
(Table~\ref{tab:b11robust}). A per-round batch effect, a noise floor fifty times the smallest spread of Appendix~\ref{app:env} and failed wells leave the
scripted policies' best number of re-tests unchanged. Under each of them, three or four re-tests a round stay below
that best (Table~\ref{tab:b11stress}). Spikes on the grid are the one exception. When 2\% of readings are three times their
value, the informed rule with three re-tests a round becomes the best scripted rule. Its pick minus that of its best
setting with fewer re-tests is \nPolSpikeHeavy{} \nPolSpikeHeavyci. Run A does not adapt to the spikes. It keeps
re-testing about \nPolShare\% of its slots, and its lead over random ordering falls to \nPolRandSpike{}
\nPolRandSpikeci{} (Table~\ref{tab:b11robust}). On GB1 and PhoQ under their real names it behaves as on the scrambled
test splits (Table~\ref{tab:b11robust}). Its lead shrinks on rougher grids and holds at every budget we tried
(Table~\ref{tab:b11robust}). At 4 rounds of 24 it is level with the best scripted rule, with two of its inputs beyond the range it saw in training.

\paragraph{The language models' own campaigns.} We replayed Run A on each model's own campaigns: the same
landscape, campaign seeds, noise and budget (Table~\ref{tab:b11llm}). We first checked that each of these landscapes
is outside the training and validation splits, by name, by built content and, for the grid, by its sorted fitness values. On
the grid Run A beats every model. On scrambled GB1 it is below GPT-6.1 Sol, and on GB1 under real names it is below
Opus 4.7.

\paragraph{What the policies learned.} We replayed every test episode and read each order back from the
environment's record (Table~\ref{tab:b11behaviour}). Run A builds from its leaders and rarely re-tests them. On the
grid, \nPolParentTopThree\% of its new candidates are one letter from one of its three highest readings. Re-tests of
those three take \nPolLeadRetest\% of its orders. Most of its re-tests retry an order that failed, \nPolRetry\%, or
repeat a candidate first read earlier the same round, \nPolSameRound\%. The seeds differ: the third seed re-tests candidates that already have a reading more often. The registered Run B re-tests from the first round, and its share climbs to about four slots in five by the
last. Most of its re-tests go to candidates it has read. On GB1 Run A's nominee is often not its highest reading, and
that costs it much of its gap to the best scripted rule there.

\begin{table}[htbp]
\centering
\small
\setlength{\tabcolsep}{1.5pt}
\caption{The registered test of the trained policy, 1{,}500 episodes per split, as a mean over three seeds. Run A trained on generated landscapes only; Run B on generated landscapes and scrambled GB1, tested on PhoQ. Differences are in final pick, with 95\% intervals from a two-level bootstrap that resamples landscapes and then episodes (post-hoc; the registered intervals resample episodes only).}
\label{tab:b11reg}
\begin{tabular}{lrrrr}
\toprule
 & Run A, grid & Run A, GB1 & Run A, PhoQ & Run B, PhoQ \\
\midrule
trained pick & 7.91 & 4.79 & 2.15 & 1.53 \\
re-testing share (\%) & 9.3 & 1.8 & 2.3 & 68.3 \\
minus initialisation & $+$0.86 [0.83, 0.89] & $+$3.32 [3.24, 3.40] & $+$1.49 [1.44, 1.54] & $+$0.87 [0.84, 0.91] \\
minus random & $+$0.40 [0.36, 0.44] & $+$1.43 [1.34, 1.53] & $+$0.65 [0.58, 0.73] & $+$0.03 [$-$0.03, 0.09] \\
minus guessing rule & $-$0.03 [$-$0.06, 0.01] & $-$1.32 [$-$1.45, $-$1.20] & $+$0.07 [$-$0.01, 0.15] & $-$0.55 [$-$0.62, $-$0.48] \\
minus informed rule & $-$0.05 [$-$0.10, 0.00] & $-$1.16 [$-$1.26, $-$1.07] & $-$0.52 [$-$0.63, $-$0.41] & $-$1.14 [$-$1.25, $-$1.03] \\
minus model-guided & $-$0.06 [$-$0.11, $-$0.01] & $-$0.56 [$-$0.67, $-$0.45] & $+$0.27 [0.21, 0.33] & $-$0.35 [$-$0.39, $-$0.31] \\
\bottomrule
\end{tabular}
\end{table}

\begin{table}[htbp]
\centering
\scriptsize
\setlength{\tabcolsep}{3pt}
\caption{The trained policy (Run A, without retraining) under every change after the registered test, 1{,}500 episodes per row, at assay noise unless the row changes it. Rows that would repeat Table~\ref{tab:b11reg} are left out. The last two columns are Run A's pick minus random ordering and minus the best scripted rule at its best number of re-tests, on the same episodes. Stress: each round's readings share one multiplicative factor with this standard deviation (batch), or get an additive noise floor of 2\% of the landscape maximum (floor). In the other two, 2\% of readings are replaced by a near-zero value (failed wells) or by three times the reading (spikes). Real names: GB1 and PhoQ under their amino-acid names. At 4 rounds of 24 the scripted rules' re-tests scale with the slots per round, and two of Run A's input features exceed the range seen in training. Intervals as in Table~\ref{tab:b11reg}; under real names, one landscape each, they resample episodes only.}
\label{tab:b11robust}
\begin{tabular}{llrrrr}
\toprule
change & condition & pick & share (\%) & minus random & minus best scripted rule \\
\midrule
noise, grid & noise halved (3.65\%) & 8.11 & 9.2 & $+$0.45 [0.40, 0.49] & $-$0.04 [$-$0.08, 0.00] \\
 & noise doubled (14.6\%) & 7.66 & 9.0 & $+$0.35 [0.31, 0.40] & $-$0.09 [$-$0.14, $-$0.05] \\
\midrule
stress, grid & batch 5\% & 7.84 & 9.3 & $+$0.41 [0.36, 0.46] & $-$0.13 [$-$0.18, $-$0.07] \\
 & batch 10\% & 7.73 & 9.2 & $+$0.40 [0.35, 0.45] & $-$0.13 [$-$0.19, $-$0.08] \\
 & floor & 7.89 & 9.3 & $+$0.40 [0.35, 0.44] & $-$0.11 [$-$0.16, $-$0.06] \\
 & failed wells & 7.90 & 9.5 & $+$0.39 [0.35, 0.43] & $-$0.10 [$-$0.14, $-$0.06] \\
 & spikes & 7.46 & 9.1 & $+$0.09 [0.05, 0.13] & $-$0.21 [$-$0.24, $-$0.17] \\
\midrule
stress, GB1 & batch 5\% & 4.78 & 1.9 & $+$1.43 [1.33, 1.53] & $-$1.23 [$-$1.35, $-$1.11] \\
 & batch 10\% & 4.73 & 1.8 & $+$1.40 [1.28, 1.51] & $-$1.22 [$-$1.35, $-$1.09] \\
 & floor & 4.82 & 1.9 & $+$1.47 [1.37, 1.58] & $-$1.26 [$-$1.38, $-$1.14] \\
 & failed wells & 4.75 & 1.8 & $+$1.42 [1.32, 1.53] & $-$1.30 [$-$1.44, $-$1.16] \\
 & spikes & 4.44 & 1.8 & $+$1.17 [1.06, 1.27] & $-$1.05 [$-$1.19, $-$0.92] \\
\midrule
real names & GB1 & 4.72 & 1.9 & $+$1.31 [1.24, 1.39] & $-$1.37 [$-$1.46, $-$1.29] \\
 & PhoQ & 2.12 & 2.3 & $+$0.58 [0.53, 0.63] & $-$0.47 [$-$0.54, $-$0.39] \\
\midrule
held-out grids & $K=2$ & 7.70 & 8.7 & $+$0.18 [0.12, 0.23] & $-$0.27 [$-$0.31, $-$0.23] \\
 & $K=3$ & 7.55 & 8.3 & $+$0.04 [$-$0.03, 0.11] & $-$0.31 [$-$0.37, $-$0.25] \\
\midrule
budgets, grid & 10 rounds of 4 & 7.83 & 9.1 & $+$0.42 [0.36, 0.47] & $-$0.06 [$-$0.12, $-$0.01] \\
 & 20 rounds of 4 & 7.97 & 9.5 & $+$0.40 [0.36, 0.44] & $-$0.11 [$-$0.15, $-$0.07] \\
 & 4 rounds of 24 & 8.01 & 11.6 & $+$0.37 [0.32, 0.43] & $-$0.04 [$-$0.09, 0.00] \\
\bottomrule
\end{tabular}
\end{table}

\begin{table}[htbp]
\centering
\scriptsize
\setlength{\tabcolsep}{3pt}
\caption{What the trained policies do on the test splits, as a mean over three seeds. Share: re-tests as a share of the budget. The next three columns split the re-tests: retries of an order that failed, repeats of a candidate first read earlier the same round, and re-tests of one of the three candidates with the highest mean reading. Parent: new candidates one letter from one of those three. Nominee: how often it is the registered selection rule's nominee, and its pick minus that nominee's.}
\label{tab:b11behaviour}
\begin{tabular}{llrrrrrrr}
\toprule
run & split & share (\%) & retries (\%) & same round (\%) & top three (\%) & parent (\%) & nominee (\%) & pick cost \\
\midrule
Run A & grid & 9.3 & 62 & 15 & 5 & 61 & 97 & $+$0.00 [0.00, 0.00] \\
Run A & GB1 & 1.8 & 58 & 29 & 4 & 63 & 51 & $-$0.71 [$-$0.75, $-$0.67] \\
Run A & PhoQ & 2.3 & 61 & 23 & 2 & 64 & 89 & $-$0.05 [$-$0.06, $-$0.04] \\
\midrule
Run B & grid & 72.0 & 3 & 5 & 58 & 43 & 71 & $-$0.03 [$-$0.04, $-$0.02] \\
Run B & GB1 & 69.5 & 3 & 5 & 42 & 35 & 93 & $-$0.06 [$-$0.07, $-$0.05] \\
Run B & PhoQ & 68.3 & 3 & 3 & 39 & 31 & 93 & $-$0.02 [$-$0.02, $-$0.01] \\
\bottomrule
\end{tabular}
\end{table}

\begin{table}[htbp]
\centering
\scriptsize
\setlength{\tabcolsep}{2pt}
\caption{Five more seeds of Run A (three to seven), trained and tested as registered. Beside them, Run B retrained with an entropy bonus five times the registered one, tested with the registered code on all three test splits. Run B trains on the grid and GB1, so only PhoQ tests its transfer. Differences in final pick against random ordering, with intervals as in Table~\ref{tab:b11reg}. Entropies are at the end of training: means over the last updates, averaged over seeds.}
\label{tab:b11phase2}
\begin{tabular}{lrrr}
\toprule
 & Run A, seeds 3 to 7 & Run B, registered & Run B, larger bonus \\
\midrule
grid: minus random & $+$0.42 [0.37, 0.46] & -- & $+$0.32 [0.28, 0.37] \\
grid: re-testing share (\%) & 10.3 & -- & 3.2 \\
grid: seeds above random & 5 of 5 & -- & 3 of 3 \\
GB1: minus random & $+$1.63 [1.53, 1.72] & -- & $+$1.66 [1.56, 1.76] \\
GB1: re-testing share (\%) & 1.7 & -- & 2.7 \\
GB1: seeds above random & 5 of 5 & -- & 3 of 3 \\
PhoQ: minus random & $+$0.66 [0.59, 0.72] & $+$0.03 [$-$0.03, 0.09] & $+$0.50 [0.43, 0.56] \\
PhoQ: re-testing share (\%) & 1.9 & 68.3 & 3.1 \\
PhoQ: seeds above random & 5 of 5 & 2 of 3 & 3 of 3 \\
order-head entropy, nats & 5.31 & 1.06 & 5.88 \\
nominate-head entropy, nats & 0.04 & 0.03 & 0.34 \\
\bottomrule
\end{tabular}
\end{table}

\begin{table}[htbp]
\centering
\scriptsize
\setlength{\tabcolsep}{3pt}
\caption{The trained policy (Run A, without retraining) on each language model's own campaigns, the same landscape, campaign seeds, noise and budget. Pick, with re-testing share; the last column is Run A's pick minus the model's, paired by campaign, with a 95\% bootstrap interval over campaigns. Each of these landscapes is outside the training and validation splits.}
\label{tab:b11llm}
\begin{tabular}{llrrrr}
\toprule
landscape & model & campaigns & model & Run A & Run A minus model \\
\midrule
grid & Opus 4.7 & 60 & 7.62 (75\%) & 7.87 (10\%) & $+$0.25 [0.14, 0.35] \\
grid & Gemini 2.5 Flash & 120 & 7.70 (94\%) & 7.86 (9\%) & $+$0.16 [0.11, 0.21] \\
grid & DeepSeek V3.1 & 120 & 7.76 (91\%) & 7.86 (9\%) & $+$0.10 [0.04, 0.15] \\
grid & GPT-6.1 Sol & 60 & 7.67 (44\%) & 7.87 (10\%) & $+$0.19 [0.07, 0.31] \\
grid & Gemini 3.1 Pro & 60 & 7.69 (14\%) & 7.87 (10\%) & $+$0.18 [0.07, 0.29] \\
\midrule
GB1, scrambled & Opus 4.7 & 60 & 4.78 (63\%) & 4.79 (2\%) & $+$0.02 [$-$0.37, 0.42] \\
GB1, scrambled & Gemini 2.5 Flash & 60 & 1.12 (92\%) & 4.79 (2\%) & $+$3.68 [3.42, 3.92] \\
GB1, scrambled & DeepSeek V3.1 & 60 & 0.84 (96\%) & 4.79 (2\%) & $+$3.95 [3.67, 4.22] \\
GB1, scrambled & GPT-6.1 Sol & 30 & 6.36 (19\%) & 4.67 (2\%) & $-$1.69 [$-$2.09, $-$1.35] \\
GB1, scrambled & Gemini 3.1 Pro & 30 & 4.90 (11\%) & 4.67 (2\%) & $-$0.24 [$-$0.82, 0.37] \\
\midrule
GB1, real names & Opus 4.7 & 60 & 7.59 (51\%) & 4.75 (2\%) & $-$2.84 [$-$3.21, $-$2.46] \\
GB1, real names & Gemini 2.5 Flash & 60 & 2.41 (45\%) & 4.75 (2\%) & $+$2.34 [1.93, 2.74] \\
GB1, real names & DeepSeek V3.1 & 60 & 1.27 (92\%) & 4.75 (2\%) & $+$3.48 [3.11, 3.84] \\
\midrule
PhoQ, scrambled & Opus 4.7 & 30 & 1.89 (63\%) & 2.20 (2\%) & $+$0.31 [0.06, 0.58] \\
PhoQ, scrambled & Gemini 2.5 Flash & 30 & 2.13 (64\%) & 2.20 (2\%) & $+$0.07 [$-$0.16, 0.33] \\
PhoQ, scrambled & DeepSeek V3.1 & 30 & 0.83 (90\%) & 2.20 (2\%) & $+$1.37 [1.17, 1.60] \\
PhoQ, scrambled & GPT-6.1 Sol & 30 & 2.37 (24\%) & 2.20 (2\%) & $-$0.17 [$-$0.44, 0.12] \\
PhoQ, scrambled & Gemini 3.1 Pro & 30 & 1.72 (15\%) & 2.20 (2\%) & $+$0.48 [0.24, 0.72] \\
\bottomrule
\end{tabular}
\end{table}

\begin{table}[htbp]
\centering
\scriptsize
\setlength{\tabcolsep}{3pt}
\caption{The guessing and informed rules under the stress conditions of Table~\ref{tab:b11robust} that keep their best number of re-tests, 1{,}500 episodes per landscape and condition, at assay noise. Without stress their best is 1 re-test a round on the grid and 0 on GB1. Best k is the best number of re-tests a round, with its re-testing share. Three minus best is the pick of three re-tests a round minus that of the best, with intervals as in Table~\ref{tab:b11reg}. Four re-tests a round are lower still in every row.}
\label{tab:b11stress}
\begin{tabular}{llrrrr}
\toprule
landscape & condition & guessing, best k & guessing, three minus best & informed, best k & informed, three minus best \\
\midrule
grid & batch 5\% & 1 (24\%) & $-$0.19 [$-$0.22, $-$0.15] & 1 (23\%) & $-$0.17 [$-$0.21, $-$0.13] \\
grid & batch 10\% & 1 (24\%) & $-$0.15 [$-$0.19, $-$0.11] & 1 (23\%) & $-$0.14 [$-$0.18, $-$0.11] \\
grid & floor & 1 (24\%) & $-$0.17 [$-$0.21, $-$0.14] & 1 (23\%) & $-$0.17 [$-$0.20, $-$0.14] \\
grid & failed wells & 1 (24\%) & $-$0.22 [$-$0.26, $-$0.17] & 1 (23\%) & $-$0.21 [$-$0.25, $-$0.17] \\
\midrule
GB1 & batch 5\% & 0 (1\%) & $-$2.21 [$-$2.34, $-$2.08] & 0 (0\%) & $-$1.58 [$-$1.70, $-$1.47] \\
GB1 & batch 10\% & 0 (1\%) & $-$2.16 [$-$2.30, $-$2.03] & 0 (0\%) & $-$1.48 [$-$1.59, $-$1.36] \\
GB1 & floor & 0 (1\%) & $-$2.27 [$-$2.41, $-$2.14] & 0 (0\%) & $-$1.68 [$-$1.81, $-$1.55] \\
GB1 & failed wells & 0 (1\%) & $-$2.32 [$-$2.46, $-$2.16] & 0 (0\%) & $-$1.71 [$-$1.82, $-$1.59] \\
\bottomrule
\end{tabular}
\end{table}

\end{document}